\documentclass{article} 
\usepackage{iclr2027_conference,times}

\usepackage{amsmath,amsfonts,bm}

\def\eqref#1{equation~\ref{#1}}

\def\1{\bm{1}}

\DeclareMathAlphabet{\mathsfit}{\encodingdefault}{\sfdefault}{m}{sl}
\SetMathAlphabet{\mathsfit}{bold}{\encodingdefault}{\sfdefault}{bx}{n}

\usepackage{hyperref}
\usepackage{url}
\usepackage[utf8]{inputenc} 
\usepackage[T1]{fontenc}    
\usepackage{hyperref}       
\usepackage{url}            
\usepackage{booktabs}       
\usepackage{amsmath}        
\usepackage{amsfonts}       
\usepackage{nicefrac}       
\usepackage{microtype}      
\usepackage{xcolor}         
\usepackage{graphicx}       
\usepackage{lipsum}
\usepackage{algorithm}
\usepackage{algpseudocode}
\usepackage{multirow}
\usepackage{enumitem}
\usepackage{wrapfig}
\usepackage{cleveref}
\usepackage{mathtools}

\title{Frontier Learning: Training LLM Reasoners at the Edge of Capability}

\newcommand{\uclaffiliation}{%
  University College London\\
  London, UK%
}

\newcommand{\baselaffiliation}{%
  University of Basel\\
  Basel, Switzerland%
}

\newcommand{\authorbox}[2]{%
  \makebox[0.42\textwidth][l]{\mdseries
    \begin{tabular}[t]{@{}l@{}}\textbf{#1}\\#2\end{tabular}}%
}

\author{%
  \authorbox{Robin Faro\footnotemark[1]}{%
    \baselaffiliation\\\texttt{robin.faro@unibas.ch}}
  \And
  \authorbox{Shyam Sundhar Ramesh\thanks{Equal contribution.}}{%
    \uclaffiliation\\\texttt{shyam.ramesh.22@ucl.ac.uk}}
  \AND
  \authorbox{Ilija Bogunovic\thanks{Equal supervision.}}{%
    \baselaffiliation}
  \And
  \authorbox{Aurelien Lucchi\footnotemark[2]}{%
    \baselaffiliation}
}

\iclrpreprintcopy
\begin{document}

\maketitle

\begin{abstract}
    Reinforcement Learning-based post-training of Large Language Models (LLM) has been successfully applied to improve their reasoning capabilities. Existing pipelines primarily finetune LLMs on a fixed pool of problems specified prior to training using the GRPO loss. This is fundamentally limiting, as learning signal arises only when policy rollouts mix successes and failures, causing the useful portion of any fixed pool to quickly become stale as the model improves. To address this, we propose \emph{frontier learning}, an open-ended post-training approach in which procedural generators are used online to continually produce informative training problems. It treats the generator's task-specific parameters as a search space and uses a regret signal to prioritize and explore frontier difficulty levels in order to focus training at the edge of the model's evolving reasoning capabilities. Across several reasoning tasks and model families, our approach consistently achieves higher relative gains over fixed-pool baselines, demonstrating that effective post-training requires not only selecting useful problems, but continually generating them at the edge of capability.
\end{abstract}

\vspace{-0.7em}
\section{Introduction}
\vspace{-0.7em}
Recent advancements in the post-training of Large Language Models (LLM) have led to drastic gains in their reasoning capabilities \citep{ouyang2022training,lightman2023let,rafailov2023direct,lambert2024tulu,guo2025deepseek}. Among various post-training techniques,
Reinforcement learning (RL) has emerged as a powerful paradigm, especially for structured reasoning domains such as math and logical reasoning~\citep{guo2025deepseek,shao2024deepseekmath}, wherein rewards are assigned in a deterministic manner using rule-based or programmatic reward functions. 
This has enabled RL-based post-training to mitigate standard pitfalls such as reward hacking and yield improved reasoning performance.\looseness=-1

Standard RL-based post-training pipelines fine-tune LLMs on a \emph{fixed pool of problems} often using Group Relative Policy Optimization (GRPO)~\citep{guo2025deepseek,shao2024deepseekmath}. However, a limitation of the GRPO objective is that it provides no optimization signal for problems whose sampled responses all receive the same reward. To address this, recent methods filter or reweight problems within the fixed pool to concentrate optimization on more informative samples~\citep{foster2025learning,yu2025dapo,bae2025online}. Such approaches are ultimately limited by the coverage of the fixed problem pool and become ineffective once the policy can reliably solve every problem in the dataset. This raises a fundamental question: \emph{How should the training
distribution adapt online to the model's evolving capability?} \looseness=-1

\begin{wrapfigure}{rt}{0.6\textwidth}
\vspace{-0.6em}
  \begin{center}
    \includegraphics[width=\linewidth]{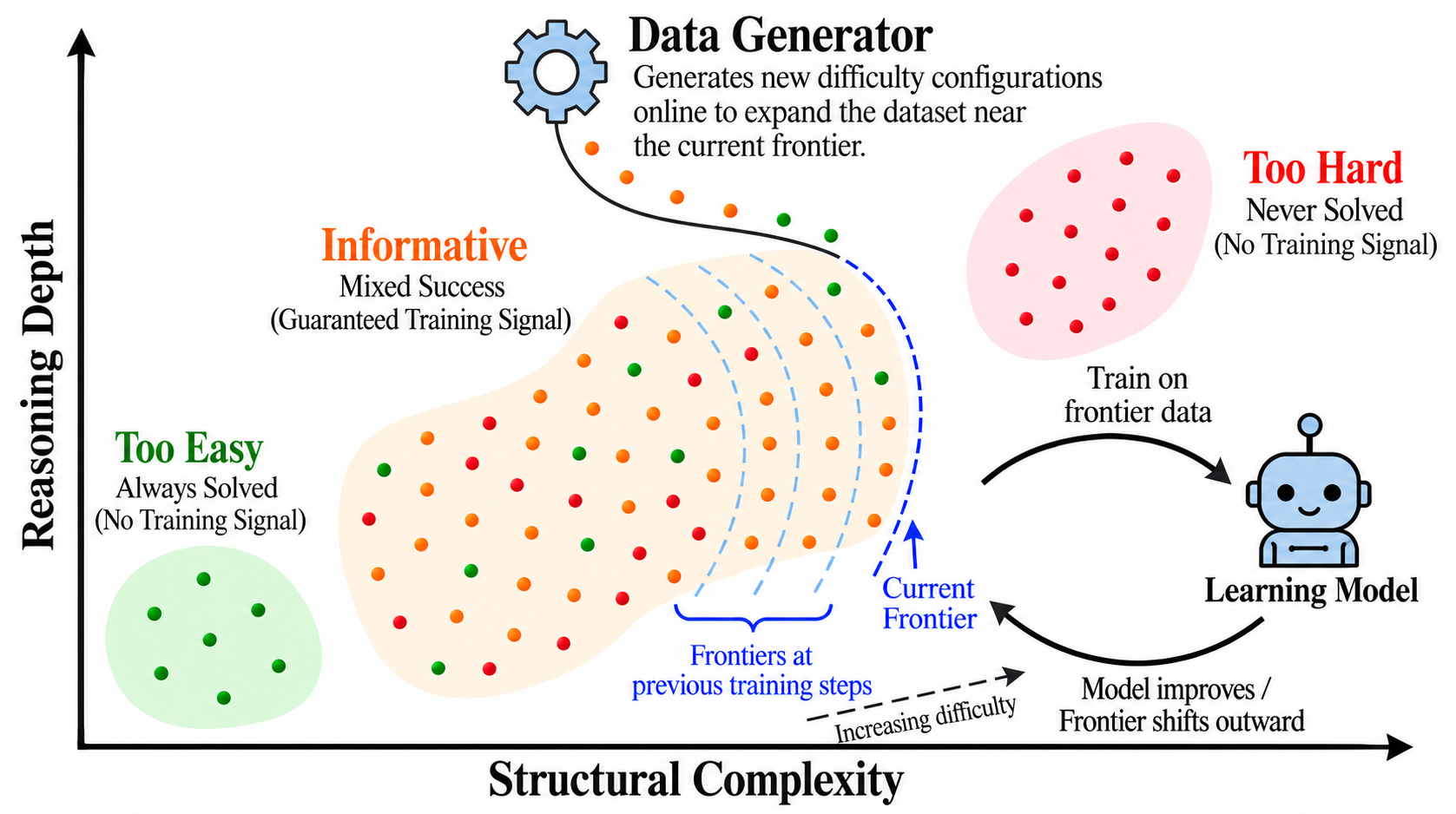}
  \end{center}
  \caption{
Frontier learning with procedural data generation. 
Levels vary along task-specific difficulty dimensions, here abstracted as structural complexity and reasoning depth. 
As the model improves, its frontier of capabilities shifts outward; frontier learning uses procedural generation to continually construct training data near this moving region.\looseness=-1
}
  \label{fig:concept-frontier}
\end{wrapfigure}

To address this, we propose to construct the training batch in an online manner using problems generated from a \emph{procedural data generator}~\citep{stojanovski2025reasoninggymreasoningenvironments,chen2025enigmata,xu2026scaler}, which are capable of synthesizing reasoning tasks at scale. In particular, procedural data generators expose task-specific attributes defining the problem structure and allow one to produce a virtually unbounded stream of training problems. However, the exposed parameters usually do not provide an explicit or monotonic measure of difficulty: different combinations can interact in complex ways, and the difficulty of a configuration is ultimately relative to the capacities of the policy. Given this, we introduce \emph{frontier learning}, an open-ended RL post-training approach that continually constructs new training data at the edge of the model's evolving capability
using the procedural data generator of a given reasoning task. Our approach actively identifies difficulty \emph{levels}, defined through the task-specific attributes, that generate \emph{problems} which are solvable enough to produce signal but not yet reliably solved by the current policy and hence, enables continual post-training using GRPO on a given reasoning task (see \Cref{fig:concept-frontier}). \looseness=-1

\textbf{Related work:}
RL-based post-training~\citep{guo2025deepseek,shao2024deepseekmath,hu2025reinforcepp,kimi2025k15}
has shown that deterministic rule-based rewards can elicit strong reasoning. However, GRPO provides no useful signal when all rollouts for a prompt are correct or incorrect~\citep{zheng2025greso,zhang2026aero,le2026rlzvp}.
Recent filtering and reweighting methods therefore concentrate training on informative prompts~\citep{foster2025learning,yu2025dapo,bae2025online,li2025knapsack,gu2026actorcurator,acegrpo2026,adacurl2025}, but remain limited by the coverage of the available problem pool.
Curriculum and learnability-based methods~\citep{bengio2009curriculum,graves2017automated,matiisen2017teacher,oudeyer2007intrinsic,willems2020mastering}
similarly prioritize examples near the learner's ability, while problem-generation methods~\citep{luo2023wizardmath,dai2026harder,sundaram2026teaching,zhao2025absolutezero}
expand the supply of training problems offline or through self-play. Frontier learning instead uses procedural generators online to explore task-specific difficulty levels, estimates level-wise regret, and mutates, explores new levels to continuously track the model's moving capability frontier.
Prior work in unsupervised environment design~\citep{dennis2020paired,jiang2021prioritized,parkerholder2022accel,campero2021amigo} has explored training in online procedural generation environments, including navigation and control tasks. In contrast, our work focuses on post-training of LLMs on reasoning tasks. We discuss related work in detail in Appendix~\ref{app:related}.\looseness=-1

\textbf{Contributions:} Our main contributions are as follows. (i) We formalize \emph{frontier learning}, an open-ended RL post-training approach that uses a procedural generator online to adapt the training curriculum to the model's evolving capability. It treats the generator's task-specific attributes as a multidimensional space of levels, without assuming that they induce a known or monotonic difficulty ordering. (ii) We design three online mechanisms that together keep the training distribution at the capability frontier: a dynamic level buffer that maintains a set of frontier levels, a regret-based priority signal that identifies and prioritizes frontier levels using only the GRPO rollouts already collected, and random exploration with state-dependent mutation that continually scouts for new frontier levels and admits them to the buffer. (iii) We instantiate frontier learning within GRPO and show that it outperforms fixed-pool, domain-randomization and adaptive-sampling baselines across five puzzle, math and graph reasoning tasks and three model families. The gains are largest over long training horizons, where frontier learning keeps improving after the baselines plateau (71.8 vs 33.3 on \textsc{Dice}), and ablations show that regret-guided mutation drives this expansion. \looseness=-1

\vspace{-0.7em}
\section{Background}
\label{sec:background}
\vspace{-0.7em}
We address the challenge of post-training a large language model (LLM), denoted by $\pi_\theta$, on a given reasoning task. This typically entails finetuning the LLM using Reinforcement Learning with Verifiable Rewards (RLVR) based on a fixed offline dataset $\mathcal{D}$, comprising problems specific to the given reasoning task, and a reward function $r(\cdot,\cdot)$. The problems admit verifiable correct answers, and the reward function $r$ is designed to evaluate the correctness of a response $y$ for any problem $x\in \mathcal{D}$. The objective in RLVR is to optimize the expected reward attained by policy $\pi_\theta$ on problems sampled from $\mathcal{D}$. Concretely, we define it as $J(\theta)=\mathbb{E}_{x\sim \mathcal{D},y\sim \pi_\theta(\cdot | x)}[r(x,y)]$.

Standard RLVR pipelines optimize $J(\theta)$ via \emph{Group
Relative Policy Optimization} (GRPO)~\citep{shao2024deepseekmath,guo2025deepseek},
which constructs advantages from within-group reward comparisons, and avoids training a
separate value network to assign the same~\citep{schulman2017proximal}. Specifically, for a given problem $x$, the policy $\pi_\theta$  generates $n$ individual responses
$\{y_1, \dots, y_{n}\}$. Then, the advantage of response $i$ given the rewards $\{r_i = r(x, y_i)\}_{i=1}^{n}$ is calculated as follows:
\begin{equation}
    A_i = \frac{r_i - \mu}{\sigma_r + \epsilon},
    \qquad
    \mu = \frac{1}{n}\sum_i r_i,\quad
    \sigma_r = \mathrm{MAD}(\{r_i\}) = \frac{1}{n}\sum_i |r_i- \mu|. 
    \label{eq:grpo-adv}
\end{equation}

Here, following \citet{dai2026harder}, we use the mean absolute deviation (MAD) in the denominator rather than standard deviation. This is to ensure the update magnitude of a single problem, determined by the sum of absolute advantages $\frac{1}{n}\sum_i |A_i|$, remains constant. This avoids disadvantaging problems at the capability frontier (all but one rollout wrong), whose absolute advantages are low when defined using the standard deviation (see \citet[Theorems-1,2]{dai2026harder}). Given the assigned advantages, GRPO optimizes the following clipped surrogate objective:
\begin{equation}
J_{\mathrm{GRPO}}(\theta)
=
\mathbb{E}_{x\sim \mathcal{D}, y_i \sim \pi_\theta(\cdot | x)}
\Bigg[
\frac{1}{n} \sum_{i=1}^n\frac{1}{|y_i|}
\sum_{t=1}^{|y_i|}
\Big(
\min\big(r_{i,t}\hat{A}_{i,t},\,
\mathrm{clip}(r_{i,t},1-\varepsilon,1+\varepsilon)\hat{A}_{i,t}\big)
\Big)
\Bigg]. 
\end{equation}

Despite the success of standard RLVR pipelines to improve the reasoning capabilities of LLMs, they still suffer from the following limitation when built upon a fixed dataset.

\paragraph{Gradient collapse:} A
well-documented failure mode of GRPO with fixed training
data~\citep{foster2025learning,bae2025online,yu2025dapo,ramesh2025multitask} is the zero gradient phenomenon wherein rollouts to a problem are all correct or all wrong. Proposed mitigations filter or resample within a fixed problem pool to increase the proportion of non-zero gradient problems in the training batch~\citep{li2025knapsack,chen2025self,wang2025dump,zhou2025daro,panaganti2026gdro,gu2026actorcurator}. Such strategies assume informative problems \emph{already exist} in the data pool and merely need to be identified from the data. The issues with such an assumption are twofold: (i) The predetermined static data might contain very low or no informative training problems, as the problems might be too easy or too hard for the given policy. (ii) Even if the dataset is designed to be informative for the policy at its initial state, as training progresses, the policy's capability improves. And once its capability reaches a certain threshold, the trained policy drifts beyond the utility of the static dataset for GRPO training. Specifically, it would have already mastered all problems in the dataset, leading to continual zero-gradient exposure and no further learning progress.  This exposes a fundamental bottleneck in GRPO post-training and points to a structural gap in standard RLVR pipelines: \emph{the training data must adapt online to the model's evolving capability.} \looseness=-1

\vspace{-0.7em}
\section{Frontier Learning}
\label{sec:fl}
\vspace{-0.7em}
In this section, we propose and formalize frontier learning, an open-ended post-training approach that allows continuous adaptation of the training data by sampling appropriate problems from procedural data generators. We begin by discussing the notion of procedural data generators upon which this approach is built.

\subsection{Procedural Data Generation}
\label{sec:proc}
Structured reasoning domains like logic, math, puzzles, etc., consist of tasks such as countdown, ARC, geometry~\citep{stojanovski2025reasoninggymreasoningenvironments} that can be generated and verified in a rule-based or programmatic manner. In such tasks, each problem is produced by an underlying procedure specified by explicit rules that samples the
terms, constraints, or values defining the problem. Such procedurally generated data has been increasingly incorporated in LLM post-training pipelines, especially to improve their general reasoning ability. \looseness=-1

Given a procedural data generator, one can specify certain task-specific attributes used by the generator to construct problems of a certain reasoning task for training. 
In particular, we define a certain attribute combination as a level $\ell \in \mathcal{L}$, where $\mathcal{L}$ is the set of all such attribute combinations, which can be intuitively regarded as a structured difficulty class from which arbitrarily many problems can be drawn. \looseness=-1

As a concrete example, consider the \textsc{Countdown} arithmetic task. A problem consists of a set of input numbers and a target number and the task is to 
produce an arithmetic expression that uses the input numbers to obtain the
target. Here, the level
$l$ will specify the number of inputs, the range of those inputs, the range of the target, and the allowed arithmetic operations. For example,
one configuration may generate puzzles with four operands sampled from
$[1,10]$ and targets in $[20,100]$, while another may generate puzzles with six
operands sampled from $[1,100]$ and targets in $[100,1000]$. Sampling
$x\sim l$ then produces one concrete puzzle from the corresponding level. Appendix~\ref{app:procedural_generator} provides an additional visual example for \textsc{Sokoban}. \looseness=-1

Although prior works incorporate data generated by such procedural generators into their pipelines, they primarily use their generative capacity only offline. They draw batches from a fixed dataset, presampled from a distribution over $\mathcal{L}$ specified prior to training. This distribution reflects the practitioner's assumptions about the model's capability but still leaves both the \emph{levels and the generated problems fixed} for the entire run. As noted in \Cref{sec:background}, such a dataset faces the same bottleneck as that of a fixed human-curated dataset for GRPO post-training, as it can only capture the informative levels in $\mathcal{L}$ according to the policy's initial capability and might fail to adapt online to the current policy's capability. \looseness=-1

\subsection{Objective for Open-ended Post-training} \label{sec:opt}
We consider the open-ended post-training framework, in which the objective is to continuously stream informative samples, w.r.t. the current policy, into the training distribution using data generated by a procedural generator. This entails actively \emph{navigating} and \emph{exploring} the set of levels $\mathcal{L}$, corresponding to the given procedural generator, 
to track the policy's evolving capability frontier. Here, the capability frontier is defined as the band of levels that comprises informative problems and contributes to learning progress. This is especially challenging, as this band of informative levels is a thin, moving slice within $\mathcal{L}$ during training. The vast majority of levels are either too easy or too difficult. Hence, sampling uniformly from the set of all levels $\mathcal{L}$ and filtering would be highly sub-optimal.

Hence, an effective approach for this framework must choose which levels to sample from at each step so as to maximize the capability of the final policy. We measure the performance of an approach tackling this framework by the following objective, evaluated at the end of $T$ training steps:
\begin{equation}
J(\theta)
= \mathbb{E}_{\ell \sim \mathcal{L},\; x \sim \ell,\;y \sim \pi_\theta(\cdot \mid x)}
\!\left[r(x, y)\right].
\label{eq:grpo_objective}
\end{equation}
Here, the objective measures the expected performance of the trained policy on instances drawn uniformly from levels $\ell \in \mathcal{L}$. In practice, we estimate $J(\theta)$ using a held-out set of levels unseen during training, thereby also assessing the agent's ability to generalize.

In order to effectively perform in the open-ended post-training framework and maximize \Cref{eq:grpo_objective}, we delineate three core requirements any approach should satisfy. First, the approach must
\emph{identify the capability frontier}. Second, it
must \emph{track this frontier over time}, maintaining training pressure
on levels that are currently near the edge of this frontier and deprioritizing those
that have become consistently solved.  Third, it must continuously \emph{expand the
training distribution}, by introducing new levels into the distribution that are not yet present
but may lie beyond the current frontier.

\subsection{Components of Frontier Learning}
\label{sec:fl}
Next, we detail our proposed frontier learning approach, aimed to tackle the open-ended post-training framework discussed in \Cref{sec:opt}.  

\begin{figure}
    \centering
    \includegraphics[width=\linewidth]{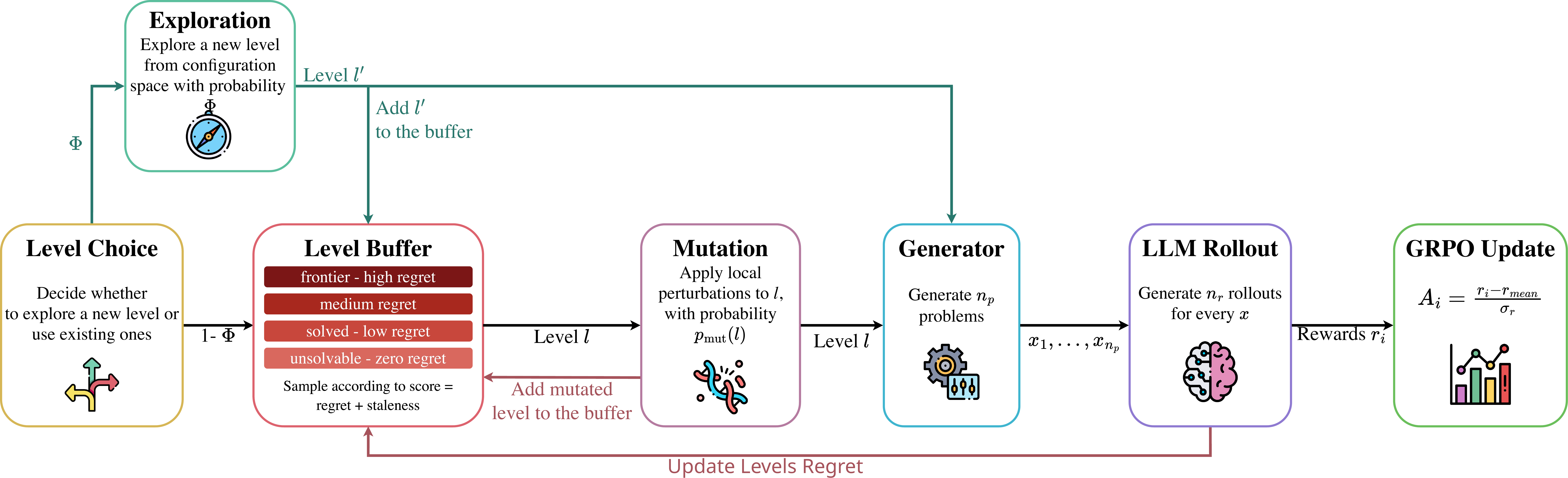}
    \vspace{-1.2em}
    \caption{
    One iteration of frontier learning with GRPO. At each step, the algorithm initially constructs the level batch with a mix of exploratory levels and high-regret frontier levels sampled from the buffer. A level sampled from the buffer is locally
    mutated with probability $p_{\mathrm{mut}}(\ell)$, by perturbing its attributes in order to evaluate and potentially extend the policy's capability frontier. The resulting level batch is then passed through the generator to construct the training batch of problems for the GRPO update. The resulting rewards and regret estimates are then used for future level selection. \looseness=-1
    }
    \label{fig:pipeline}
    \vspace{-1.5em}
\end{figure}

Frontier learning addresses the core requirements detailed in \Cref{sec:opt} through three tightly coupled online
mechanisms operating on a dynamic \emph{level buffer} $\mathcal{B}$, a
fixed-capacity set of levels continuously replenished and pruned
throughout training. In particular, it uses a \emph{regret-based priority signal} to identify
levels with remaining learning potential and concentrates sampling on
the current frontier.  Moreover, it conducts \emph{exploration} to continuously scout $\mathcal{L}$ for informative levels that exist beyond the buffer.  Finally, it also performs \emph{state-dependent mutation}
to perturb the attributes of frontier levels and incorporate levels at incrementally harder
difficulties into the level buffer $\mathcal{B}$. We summarize our approach in \Cref{fig:pipeline}. The following subsections define each mechanism in turn.

\subsubsection{Regret and Priority}
\label{sec:regret}
We define regret at two levels of granularity: per \emph{problem} $x$ and per \emph{level} $\ell$.

\textbf{Problem-wise regret:}
For a problem $x$ drawn from level $\ell$, the policy generates $n$ rollouts with $s$ successes.
Two quantities characterise the policy's behaviour on $x$: the \emph{solvability indicator} $\mathbf{1}[s \geq 1]$, which is $1$ if the policy solves $x$ at all and $0$ otherwise, and the \emph{empirical success rate} $s(x) = s / n$.
The problem-level regret is their difference:
\begin{equation}
    \rho(x)
    = \mathbf{1}[s \geq 1] - s(x).
    \label{eq:inst-regret}
\end{equation}
Regret is zero for impossible problems ($s=0$, no gradient signal exists) and for fully mastered ones ($s(x)=1$, solvability and success rate coincide), and is strictly positive only in the productive regime where gradient signal flows.

\textbf{Level-wise regret:}
The regret of a level $\ell$ is the expected problem-wise regret over problems drawn from its generator:
\begin{equation}
    \mathcal{R}(\ell)
   = \mathbb{E}_{x \sim \ell}\bigl[\rho(x)\bigr]
    = \mathbb{E}_{x \sim \ell}\!\Bigl[
        \mathbf{1}[s \geq 1] - s(x)
      \Bigr].
    \label{eq:regret}
\end{equation}

Crucially, a level accumulates high regret if and only if many of its problems are solvable but not yet reliably solved, i.e., many problems lie in the informative band. A level where most problems are either impossible or already mastered yields low $\mathcal{R}(\ell)$ regardless of a few outliers, correctly reflecting limited learning potential. \looseness=-1

In practice, we estimate $\mathcal{R}(\ell)$ by tracking the moving window of problems $\mathcal{W}_l$ sampled from $\ell$:
\begin{equation}
    \hat{\rho}(\ell)
    = \frac{1}{|\mathcal{W}_\ell|} \sum_{x \in \mathcal{W}_\ell} \rho(x).
    \label{eq:regret-est}
\end{equation}
Here, $\rho(x)$ is computed directly from the $n$ rollouts collected on $x$ during the rollout phase, and hence no additional rollouts are required. 
As the policy improves across problems from $\ell$, $\hat{\rho}(\ell)$ falls automatically, and the buffer starts prioritizing harder levels that have newly entered the informative band. \looseness=-1

\textbf{Priority score:}
Each level is assigned a priority score that adds a staleness
correction to regret:
\begin{equation}
    P(\ell, t)
    = \mathcal{R}(\ell)
    + \lambda_s \cdot (t - t_\ell),
    \label{eq:priority}
\end{equation}
where $t_\ell$ is the last step at which level $\ell$ was sampled.  The staleness
term ensures all levels in the buffer are periodically revisited and helps detect changes in the policy's capability on dormant levels.\looseness=-1

\subsubsection{Regret-Guided Sampling with Exploration}
\label{sec:replay}

At each training step $t$, a batch of $n_\ell$ levels is assembled as follows:
All $n_\ell$ slots are initially drawn from the dynamic level buffer
$\mathcal{B}$ by a Zipfian distribution over priority scores $P(\ell, t)$,
concentrating training on the current capability frontier while
preserving diversity across the buffer.  Each slot is then
independently replaced with probability $\phi$ with a random level sampled uniformly
from $\mathcal{L}$. This ensures that the expected fraction of exploratory levels in the current training batch is $\phi$.  Moreover, this helps the training batch to stay anchored at the productive frontier, while also exploring continuously for more difficult levels beyond the current buffer to avoid stagnation. \looseness=-1


\subsubsection{State-Dependent Mutation}
\label{sec:mutation}
It is important to expand the buffer with frontier levels at incrementally harder difficulties so that it remains continually informative. However, neither regret nor exploration systematically target the boundary of the model's current capability. Mutation fills this gap by applying a small, undirected perturbation to existing buffer levels to produce new levels in their local neighbourhood.

At every step $t$, each level $\ell$ that was sampled from the buffer and not replaced with a random level due to exploration is mutated with a probability that depends on its current
difficulty class defined w.r.t. the average empirical success rate of its problems $\hat{s}(\ell)
    = \frac{1}{|\mathcal{W}|} \sum_{x \in \mathcal{W}_\ell} s(x)$ in the last $\mathcal{W}$ :
\vspace{-0.7em}

{\footnotesize
\begin{equation}
    p_{\mathrm{mut}}(\ell) =
    \begin{dcases*}
        p_{\mathrm{inf}} & $\hat{s}(\ell) \in (0.05, 0.95)$ (informative) \\
        p_{\mathrm{easy}} & $\hat{s}(\ell) > 0.95$ (too easy) \\
        p_{\mathrm{hard}} & $\hat{s}(\ell) < 0.05$ (too difficult)
    \end{dcases*}
    \label{eq:mutprob}
\end{equation}}
with $p_{\mathrm{inf}} > p_{\mathrm{easy}} \geq p_{\mathrm{hard}}$.  This
concentrates mutation effort on informative levels, while still allowing easy levels to potentially produce new harder variants and enabling impossible levels to produce easier variants in a limited manner so as not to populate the buffer with a lot of impossible levels. \looseness=-1

The combined effect of state-dependent mutation probabilities is that the buffer evolves to track the model's improving capability as informative frontier levels are continuously allowed to seed variants at incrementally higher difficulties.  
\vspace{-0.7em}
\section{Algorithm}
\vspace{-0.7em}
In this section, we present an instantiation of our frontier learning approach with GRPO in \Cref{alg:accel_grpo}. In particular, \Cref{alg:accel_grpo} comprises of three tightly coupled operations at each step, namely, assembling a regret-guided training batch using a procedural data generator, updating the policy via GRPO, and refreshing the regret state using the newly observed returns (illustrated in \Cref{fig:pipeline}). We describe the main components of \Cref{alg:accel_grpo} below.

\begin{algorithm}[t]
\caption{Frontier Learning with GRPO}
\label{alg:accel_grpo}
\begin{algorithmic}[1]
\Require Level space $\mathcal{L}$, reward $r$, policy $\pi_\theta$,
buffer capacity $B_{\max}$, initial regret $\mathcal{R}_0$, explore fraction $\phi$,
staleness coeff. $\lambda_s$, mutation probs. $p_{\mathrm{inf}},p_{\mathrm{easy}},p_{\mathrm{hard}}$
\State Initialise $\mathcal{B}$ by grid seeding $\mathcal{L}$; set
$\mathcal{R}(\ell)\!\leftarrow\!\mathcal{R}_0$ and $P(\ell,0)\!\leftarrow\!\mathcal{R}(\ell)$ for all $\ell\in\mathcal{B}$
\For{$t=1,\dots,T$}
    \State Sample $n_\ell$ levels from $\mathcal{B}$ by Zipfian sampling over priority scores $P(\ell,t)$
    \State $\mathcal{L}_{\mathrm{new}}\leftarrow\emptyset$
    \For{each sampled $\ell$}
        \State With prob. $\phi$, draw $\ell'\sim\mathcal{L}$; if $\ell'\notin\mathcal{B}$, admit $\ell'$, add it to $\mathcal{L}_{\mathrm{new}}$, replace $\ell\leftarrow\ell'$ \Comment{Explore}
    \EndFor
    \For{each sampled $\ell\notin\mathcal{L}_{\mathrm{new}}$}
        \State With prob. $p_{\mathrm{mut}}(\ell)$, perturb $\ell\to\ell'$; if $\ell'\notin\mathcal{B}$, admit $\ell'$, replace $\ell\leftarrow\ell'$ \Comment{Mutate}
    \EndFor
    \For{each sampled $\ell$}
        \State Draw $n_p$ problems with $n_r$ rollouts each: $x\sim \ell$, $\{y_i\sim\pi_\theta(\cdot\mid x)\}_{i=1}^{n_r}$.
        \State Compute $r_i=r(x,y_i)$
    \EndFor
    \State Update $\theta$ with GRPO on $\{(x,y_i,r_i)\}$ \Comment{Eq.~\ref{eq:grpo-adv}}
    \For{each sampled $\ell$}
        \State Append rewards to $\ell$'s rolling window; update $\mathcal{R}(\ell)$ and $P(\ell,t)$ \Comment{Eqs.~\ref{eq:inst-regret}--\ref{eq:priority}}
    \EndFor
\EndFor
\State \Return $\pi_\theta$
\end{algorithmic}
\end{algorithm}

\textbf{Grid seeding initialisation (line~1):} The buffer is populated by \emph{grid seeding}, wherein $\mathcal{L}$ is partitioned
along each attribute axis into an equal number of cells and one level is sampled uniformly from each cell.  This guarantees broad coverage of the difficulty spectrum from the very first step, preventing the cold-start collapse that occurs when random seeding places all initial levels in a single difficulty band.

\textbf{Batch assembly (lines~3--10):}
The batch is assembled from $n_\ell$ replay levels drawn from $\mathcal{B}$ by Zipfian sampling over priority scores $P(\ell,t)$ (see \Cref{eq:priority}). Then, each sampled level is independently considered for exploration: with probability $\phi$, we draw a candidate $\ell' \sim \mathrm{Uniform}(\mathcal{L})$, check whether $\ell' \notin \mathcal{B}$, and, if novel, admitted to the buffer $\mathcal{B}$ and used in place of the original replay level.  Then, each non-replaced replay level is offered to the mutator: with probability $p_\mathrm{mut}(\ell)$ (\Cref{eq:mutprob}), a single attribute is perturbed to produce a candidate child level $l'$. If $l'$ is novel, it is admitted to the buffer $\mathcal{B}$ and used in place of the original replay level. The constructed batch is then passed on to the procedural generator. When $|B|=B_{\max}$, admitting a new level evicts the level with the lowest priority score $P(\ell,t)$. \looseness=-1

\textbf{Rollout and policy update (lines~11--15):}
For each active level $\ell$ in the batch, $n_p$ problems
$x \sim \ell$ are generated using the procedural generator and $n$ rollouts are sampled from the current
policy for each problem $x$. The full batch is then used to update $\theta$ via GRPO
(\Cref{eq:grpo-adv}). 

\textbf{Buffer update (lines~16--18):}
After the policy update, returns from every trained level are appended to its
rolling window and regret $\mathcal{R}(\ell)$ and priority $P(\ell,t)$ are recomputed from \Cref{eq:regret,eq:priority}.  Hence, the buffer's priority scores constantly reflects the \emph{updated} policy's capability frontier, and levels with low regret are immediately deprioritised and the priority shifts towards the new frontier levels with high regret at the next batch assembly.
\vspace{-0.7em}
\section{Experiments}
\label{sec:experiments}
\vspace{-0.7em}
We evaluate our proposed frontier learning with GRPO (\Cref{alg:accel_grpo}) on five procedurally generated reasoning tasks drawn from Reasoning
Gym~\citep{stojanovski2025reasoninggymreasoningenvironments}, each parameterised in a structured level space $\mathcal{L}$, spanning three domains.
In the \emph{puzzle} domain, \textsc{Countdown} requires combining numbers via basic arithmetic to reach a target (difficulty grows with operands count, their magnitude, and the
target range), and \textsc{Sokoban} requires planning box-pushing moves on a grid (harder problems have larger grids, more boxes, and deeper solution traces).
Concerning the \emph{math} domain, we include \textsc{Dice}, a discrete probability task in which the model must compute the exact probability of a target outcome when rolling a set of
multi-sided dice (where difficulty is controlled by the number of dice and their face count),
and \textsc{Decimal Arithmetic}, which consists of fixed-precision
multi-term decimal calculations, scaling in difficulty with term count, decimal places, and required precision.
Finally, in the \emph{graph} domain, \textsc{Largest Island} requires the model to identify the largest connected component of 1-valued cells in a binary grid and report its area. Difficulty scales with grid dimensions and the number of islands, requiring the model to trace connected components over increasingly large graphs. \looseness=-1

\textbf{Models and training protocol:} We experiment with two model variants from the Qwen3 family~\citep{qwen3}, chosen
to explore the effectiveness of frontier learning across two different
starting capability regimes. The pre-trained only \textbf{Qwen3-4B-Base} is used for \textsc{Countdown}, \textsc{Dice}, and \textsc{Decimal Arithmetic}. For \textsc{Sokoban}, we use the instruction-tuned \textbf{Qwen3-4B} with thinking mode enabled, since such puzzles require more sustained multi-step reasoning and therefore benefit from a larger reasoning budget. To assess whether the effectiveness of frontier learning generalizes across model families and model sizes, we additionally conduct experiments on \textsc{Countdown} using \textbf{Llama-3.2-3B-Instruct}~\citep{grattafiori2024llama3herdmodels} and on \textsc{Largest Island} using \textbf{Olmo3-7B-Instruct}~\citep{olmo2025olmo3}.
For \textsc{Countdown}, \textsc{Sokoban}, and \textsc{Decimal Arithmetic}, models are trained for $T = 200$ GRPO steps, while for \textsc{Dice} and \textsc{Largest Island} we extend the training horizon to $T=500$ GRPO steps, in order to analyze how our method behaves under prolonged training. Full implementation details and hyperparameters are reported in Appendix~\ref{app:hyperparameters}, while a detailed accounting of the used compute is provided in Appendix~\ref{app:compute}.

\textbf{Baselines:} We compare frontier learning against four representative alternatives for selecting
training levels from procedural generators. Three of them rely on a fixed level buffer seeded once before training via grid seeding,  introducing no new levels during training, and hence isolating the effect of having a fixed pool of levels in contrast to \emph{growing} it.
In particular, \textbf{Uniform} samples uniformly from the seeded buffer without any priority, providing a clean upper bound on what structured initial coverage alone can achieve. Both \textbf{SEC}~\citep{chen2025self} and \textbf{PLR}~\citep{jiang2021prioritized} attempt instead to recreate a curriculum within the fixed initial buffer by assigning priority scores to levels and sampling more frequently from those estimated to be most productive.
SEC estimates productivity by using the absolute GRPO advantage estimated from previous rollouts, while PLR uses the same level-wise regret signal $\mathcal{R}(\ell)$ as frontier learning (Eq.~\ref{eq:regret}).
On the other hand, \textbf{Domain Randomization (DR)} samples fresh levels at every step from the same coverage-aware seeding grid used to initialize the other baselines, with no memory or replay, thereby testing unguided exploration of the level space. \looseness=-1

\textbf{Evaluation:} For each task, we pre-define a fixed set of evaluation levels $\ell_v \sim \mathcal{L}$, each associated with a total of 200 unique problems $x_{val}$. Each level's configuration is excluded from grid initialization, uniform exploration, and mutation admission. Thus, neither the evaluation instances nor their exact generator configurations are encountered during training. Levels are ordered by a task-specific difficulty heuristic derived from their configuration attributes, yielding a spectrum that we partition into three bins (Easy, Medium, Hard); full definitions and the bin partition are
given in Appendix~\ref{app:anchor_levels}. The primary metric is \textbf{accuracy}: the mean greedy-decoding accuracy across all evaluation levels, averaged over four seeds.\looseness=-1

\subsection{Results}
\label{sec:results}
\textbf{Performance under the standard training budget:} We first evaluate frontier learning on \textsc{Countdown}, \textsc{Decimal Arithmetic}, and
\textsc{Sokoban} using the standard budget of 200 GRPO steps.
As shown in Table~\ref{tab:main_results}, frontier learning achieves the highest
average final performance on all three tasks. 

\begin{wraptable}[12]{r}{0.55\textwidth}
\centering
\vspace{-0.73cm}
\fontsize{8}{9}\selectfont
\caption{
Accuracy (\%) on the fixed anchor evaluation set after 200 training steps.
Best result per task in \textbf{bold}; second-best \underline{underlined}.
}
\label{tab:main_results}
\setlength{\tabcolsep}{1pt}
\renewcommand{\arraystretch}{0.8}
\begin{tabular*}{\linewidth}{@{\extracolsep{\fill}} l c c c @{}}
\toprule
& \multicolumn{2}{c}{\textit{Puzzle}} & \textit{Math} \\
\cmidrule(lr){2-3}\cmidrule{4-4}
\textit{Task} & Countdown & Sokoban & Dec.\ Arith. \\
\midrule
\textit{Model} & \textit{Qwen3-4B-Base} & \textit{Qwen3-4B} & \textit{Qwen3-4B-Base} \\
\midrule
\textit{Method} & & & \\
Domain Rand.      & \underline{44.7}$_{\pm 1.7}$ & 37.6$_{\pm 2.6}$          & \underline{31.0}$_{\pm 2.8}$ \\
Uniform           & 44.2$_{\pm 4.0}$             & 39.5$_{\pm 1.3}$          & 28.0$_{\pm 2.6}$ \\
SEC               & 40.1$_{\pm 8.1}$             & 34.3$_{\pm 6.6}$          & 29.1$_{\pm 4.1}$ \\
PLR               & 44.3$_{\pm 2.9}$             & \underline{41.1}$_{\pm 3.9}$ & 27.5$_{\pm 2.9}$ \\
Frontier Learning (ours) & \textbf{50.9}$_{\pm 0.5}$ & \textbf{45.2}$_{\pm 1.9}$ & \textbf{34.7}$_{\pm 3.6}$ \\
\midrule
Relative gain vs 2nd (\%) & \textbf{+13.9} & \textbf{+10.0} & \textbf{+11.9} \\
\bottomrule
\end{tabular*}
\end{wraptable}
The clearest improvement is observed
on \textsc{Countdown}, where frontier learning reaches
$50.9\pm0.5$\% accuracy, compared with $44.7\pm1.7$ for the strongest baseline, while the overall gains are smaller on \textsc{Sokoban} and \textsc{Decimal Arithmetic}.

The per-difficulty results, shown in Figure \ref{fig:level_bins_bars}, provide a more detailed view of these differences. On \textsc{Sokoban}, frontier
learning's largest gain is concentrated on easier configurations that are not yet
fully mastered, while more advanced levels remain largely unproductive. On
\textsc{Decimal Arithmetic}, the improvement is concentrated on Easy and Medium
levels, whereas all methods remain effectively tied on Hard configurations.
These results show that frontier learning successfully identifies where the model has the greatest remaining room for improvement and concentrates training on those configurations. Appendix~\ref{app:learning_curves} reports the full learning curves for all methods on these tasks.

\begin{wraptable}[15]{r}{0.42\textwidth}
\centering
\vspace{-0.8cm}
\fontsize{8}{9}\selectfont
\caption{
Accuracy (\%) on \textsc{Dice} for the baselines, Frontier Learning and its components ablation.
Best result in \textbf{bold}; second-best within each block \underline{underlined}.
}
\label{tab:dice_500}
\setlength{\tabcolsep}{1.2pt}
\renewcommand{\arraystretch}{0.8}
\begin{tabular*}{\linewidth}{@{\extracolsep{\fill}} l c @{}}
\toprule
Method & Accuracy \\
\midrule
\addlinespace[1pt]
Domain Rand.   & 23.4$_{\pm 4.7}$ \\
Uniform        & 31.9$_{\pm 1.5}$ \\
SEC            & \underline{33.3}$_{\pm 1.2}$ \\
PLR            & 31.2$_{\pm 1.9}$ \\
ACE-GRPO       & 33.2$_{\pm 2.1}$ \\
DAPO           & 25.4$_{\pm 5.9}$ \\
Frontier Learning (ours) & \textbf{71.8}$_{\pm 6.2}$ \\
\midrule
\multicolumn{2}{@{}l}{\textit{Component ablations}} \\
\addlinespace[1pt]
Unif+Explore          & 30.8$_{\pm 1.3}$ \\
Unif+FlatMut          & 47.1$_{\pm 10.8}$ \\
Unif+Explore+FlatMut  & 36.9$_{\pm 2.9}$ \\
PLR+Explore           & 33.5$_{\pm 1.8}$ \\
PLR+Mut               & \underline{62.5}$_{\pm 4.5}$ \\
\midrule
Relative gain vs best baseline (\%) & \textbf{+115.6}
\\
\bottomrule
\end{tabular*}
\end{wraptable}
\textbf{Long-horizon frontier expansion on \textsc{Dice}:} We next extend training on \textsc{Dice} from 200 to 500 GRPO steps to test
whether frontier learning continues to track the model's improving capabilities
over a longer horizon. In addition to the baselines introduced above, we include DAPO-style dynamic-filtering baseline~\citep{yu2025dapo} and an ACE-GRPO sampling baseline~\citep{acegrpo2026}. DAPO discards
rollout groups whose rewards are uniformly correct or uniformly incorrect.
Because this filtering can consume multiple rollout groups for each optimizer
update, we match DAPO to the total rollout budget of the other methods,
corresponding to 256k rollouts, rather than to 500 optimizer updates.  ACE-GRPO instead prioritizes replay-buffer levels using its Learnability Potential score. \looseness=-1

Figure~\ref{fig:dice_learning_curves} shows that the standard baselines improve
during the first approximately 100--200 steps and subsequently plateau. Frontier
learning instead continues improving throughout the longer training horizon,
reaching $71.8\pm6.2$\% accuracy after 500 steps. The strongest standard baseline,
SEC, reaches $33.3\pm1.2$, while ACE-GRPO reaches $33.2\pm2.1$ and the
rollout-matched DAPO baseline reaches $25.4\pm5.9$. 

Additionally, Figure~\ref{fig:dice-final-bin-difficulty}  separates performance by
difficulty. Frontier learning reaches near-perfect performance on Easy and Medium
configurations while making substantial progress on regions that remain almost
entirely unsolved by the baselines. In particular, it reaches
$72.6\%$ on Hard levels and $20.7\%$ on Extra-hard levels while all the
standard baselines remain under $5.5\%$ on such categories. The long-horizon improvement is therefore
not limited to further optimizing already-mastered configurations, but reflects
continued expansion of the model's capability frontier. 
Appendix~\ref{app:difficulty_breakdown} provides the per-difficulty learning curves, showing how performance across the individual difficulty bins evolves throughout training.  We reuse this same experimental setup to run a sweep across step sizes for two baselines, reported in Appendix~\ref{app:lr_sweep}. \looseness=-1

\textbf{Component ablation:}
We ablate buffer growth, random exploration, mutation, and regret-based prioritization under the same 500-step \textsc{Dice} protocol. All ablations use a dynamic buffer. \textbf{PLR+Explore} removes mutation, \textbf{PLR+Mut} removes random exploration, the Uniform variants remove regret-based prioritization, and \textbf{FlatMut} replaces state-dependent mutation with a level-independent probability.

Table~\ref{tab:dice_500} shows that buffer growth alone is insufficient: PLR+Explore reaches only $33.5\pm1.8$. By contrast, PLR+Mut reaches $62.5\pm4.5$, showing that mutation-driven expansion is key to moving beyond the seeded difficulty range. Mutation alone is also insufficient: Unif+FlatMut reaches $47.1\pm10.8$, substantially below PLR+Mut, indicating that regret-guided prioritization makes mutation more effective by focusing replay and subsequent mutations on informative levels. The full method reaches $71.8\pm6.2$, with random exploration providing a further gain, especially on Hard and Extra-hard levels.
\looseness=-1

\begin{figure}[h]
    \centering
    \includegraphics[width=0.75\linewidth]{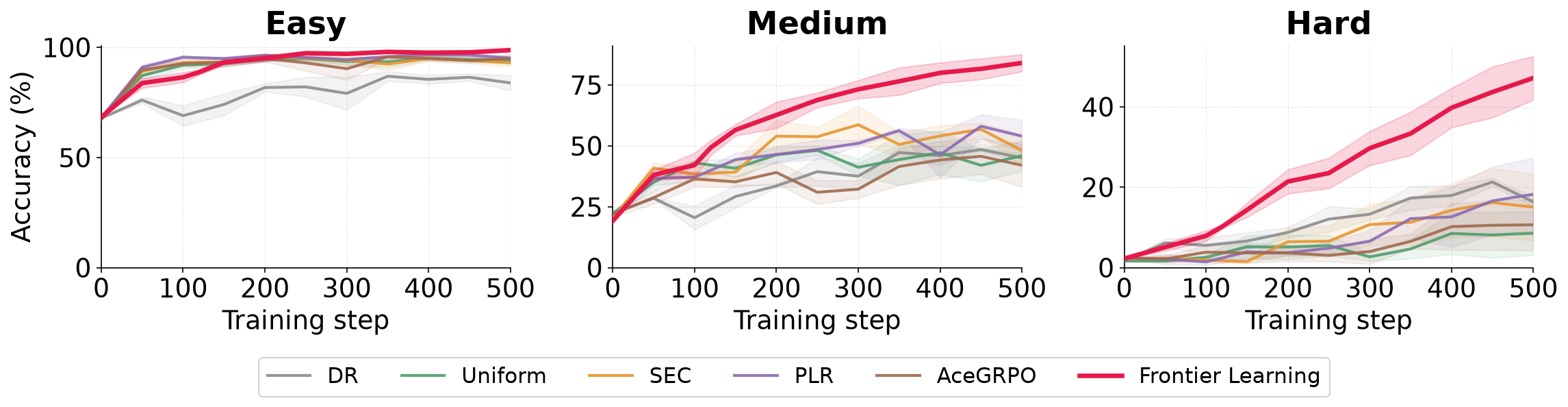}
    \caption{
Per-difficulty accuracy (\%) on \textsc{Largest Island} with Olmo3-7B-Instruct
over 500 GRPO steps.
All methods saturate on Easy levels, but only frontier learning keeps improving
on Medium and Hard levels throughout training, while the baselines plateau
after roughly 200 steps.
}
    \label{fig:largest_island_curves}
\end{figure}

\begin{wraptable}[12]{r}{0.55\textwidth}
\centering
\vspace{-0.8cm}
\fontsize{8}{9}\selectfont
\caption{
Accuracy (\%) on \textsc{Countdown} and
\textsc{Largest Island}.
Best result per task in \textbf{bold}; second-best \underline{underlined}.
}
\label{tab:generalization}
\setlength{\tabcolsep}{1pt}
\renewcommand{\arraystretch}{0.8}
\begin{tabular*}{\linewidth}{@{\extracolsep{\fill}} l c c @{}}
\toprule
\textit{Task}  & Countdown & Largest Island \\
\midrule
\textit{Model} & Llama-3.2-3B-Instruct & Olmo3-7B-Instruct \\
\midrule
\multicolumn{3}{@{}l}{\textit{Method}} \\
\addlinespace[1pt]
Domain Rand. & \underline{33.8}$_{\pm 1.9}$ & 45.2$_{\pm 1.8}$ \\
Uniform      & 25.3$_{\pm 4.7}$ & 45.3$_{\pm 3.4}$ \\
SEC          & 29.5$_{\pm 1.6}$ & 48.3$_{\pm 4.7}$ \\
PLR          & 29.0$_{\pm 1.8}$ & \underline{52.0}$_{\pm 5.3}$ \\
ACE-GRPO     & 31.4$_{\pm 1.9}$ & 45.3$_{\pm 5.5}$ \\
Frontier learning (ours) & \textbf{39.2}$_{\pm 1.3}$ & \textbf{73.7}$_{\pm 3.3}$ \\
\midrule
Relative gain vs 2nd (\%) & \textbf{+16.0} & \textbf{+41.7} \\
\bottomrule
\end{tabular*}
\end{wraptable}
\textbf{Generalization across model families and sizes:} To assess whether frontier learning remains effective outside the
Qwen3 family and the 4B model size, we train \textbf{Llama-3.2-3B-Instruct} on \textsc{Countdown} under the standard 200-step budget and \textbf{Olmo3-7B-Instruct} on
\textsc{Largest Island} under the extended 500-step budget.
Table~\ref{tab:generalization} shows that frontier learning attains the best
accuracy in both settings.
The per-difficulty learning curves for \textsc{Largest Island}
(Figure~\ref{fig:largest_island_curves}) mirror the \textsc{Dice} results shown in Figure~\ref{fig:level_bin_curve_dice}:
all methods saturate on Easy levels, but on Medium and Hard levels the
baselines plateau after roughly 200 steps while frontier learning continues to
improve until the end of training, reaching $84.0\%$ and $47.2\%$ respectively
against at most $54.0\%$ and $18.2\%$ for any baseline. These results indicate that the gains of frontier
learning are not tied to a particular model family, scale, or task domain.
We additionally use the Llama \textsc{Countdown} setting for the hyperparameter sensitivity
analysis in Appendix~\ref{app:hyperparameter_sensitivity}.
\looseness-1

\vspace{-0.8em}

\clearpage
\vspace{-0.7em}
\section{Conclusions}
\label{sec:conclusion}
 \vspace{-0.7em}
We formalize frontier learning, a novel open-ended RL post-training approach that adapts the training curricula for finetuning LLMs on reasoning tasks using a procedural generator. We design three online mechanisms for our approach, namely, a dynamic level buffer to maintain a set of frontier levels, a regret-based priority signal to identify and prioritize the frontier levels and exploration, mutation to constantly scout and incorporate new frontier levels into the buffer. Together, they ensure the training distribution is constantly adapted online to the model's evolving capability. We instantiate our approach within the GRPO objective and empirically demonstrate that our approach consistently outperforms baselines on several puzzle, math and graph tasks.

Our approach is limited to a procedural generator equipped with continuous or discrete task-specific parameters. Extending frontier learning to tasks where problem instances cannot be procedurally generated remains an interesting avenue for future work.

\bibliography{iclr2027_conference}
\bibliographystyle{iclr2027_conference}
\clearpage
\appendix

\section{Additional Related Work}

\label{app:related}

\paragraph{Adaptive sampling and curricula in RLVR.}

The zero-gradient failure mode of GRPO has motivated methods that concentrate rollout compute on informative problems~\citep{zheng2025greso,zhang2026aero,le2026rlzvp,xiong2025reinforceada}. Sampling for Learnability~\citep{foster2025learning} and DAPO~\citep{yu2025dapo} prioritise prompts with mixed rollout outcomes, with theoretical support for targeting intermediate-difficulty problems provided by \citet{bae2025online}. Other approaches adapt rollout allocation, task sampling, or difficulty-group weighting~\citep{li2025knapsack,ramesh2025multitask,chen2025self,wang2025dump,zhou2025daro,panaganti2026gdro}.  Working with a fixed problem pool, AdaCuRL schedules difficulty buckets with historical revisitation~\citep{adacurl2025}, while Actor-Curator trains a neural curator to select problems according to estimated policy improvement~\citep{gu2026actorcurator}. AceGRPO additionally constructs and prioritises training tasks from intermediate execution states~\citep{acegrpo2026}.  Frontier learning instead searches a procedural generator's configuration space, using configuration-level regret to determine where new training problems should be generated.

\paragraph{Adaptive procedural generation.}
Procedural reasoning frameworks such as Reasoning Gym~\citep{stojanovski2025reasoninggymreasoningenvironments}, SynLogic~\citep{liu2025synlogic}, and Reasoning Core~\citep{lacombe2026reasoningcore} provide scalable generators of verifiable reasoning problems with controllable task parameters or difficulty, enabling training distributions beyond fixed datasets. Recent work has begun adapting such distributions to the evolving policy.  SCALER~\citep{xu2026scaler} generates problems at difficulty levels that it raises or lowers according to the model's current accuracy. This relies on a predefined ordering of difficulty levels. Frontier learning considers a different setting in which the generator exposes a multidimensional configuration space whose parameters need not induce a known or monotonic difficulty ordering.

\paragraph{Unsupervised environment design and learnability.}

Frontier learning is also related to unsupervised environment design (UED), where the training environment evolves with the policy. PAIRED generates environments according to protagonist--antagonist regret~\citep{dennis2020paired}, while PLR prioritises previously encountered levels according to learning-potential estimates~\citep{jiang2021prioritized}. More closely related algorithmically, ACCEL~\citep{parkerholder2022accel} mutates high-regret levels to generate new environments near the agent's evolving capability frontier.  Frontier learning builds on this replay-and-editing principle in RLVR, estimating regret from groups of binary, verifiable rollout outcomes and aggregating it across problems from each procedural configuration. These configuration-level statistics guide replay alongside state-dependent mutation and exploration of unseen configurations. This signal is also related to learnability-based curricula~ \citep{oudeyer2007intrinsic,graves2017automated,matiisen2017teacher,foster2025learning,chen2025self}; unlike symmetric mixed-outcome criteria such as the binary-reward SEC score, our regret prioritises the outer edge of the productive region and uses it not only for sampling but also to determine where the configuration space should expand.

\paragraph{LLM-based problem and environment generation.}
A complementary line of work uses LLMs themselves to generate or transform training problems~\citep{luo2023wizardmath,dai2026harder,hosseini2024vstar,guan2025rstarmath,sundaram2026teaching}.  Several recent methods couple generation directly to the learner's evolving capabilities. CLPO uses the learner itself to simplify hard problems and diversify intermediate-difficulty problems through answer-preserving rewriting~\citep{zhang2026clpo}. Absolute Zero~\citep{zhao2025absolutezero} jointly learns to propose and solve tasks through self-play, while R-Zero~\citep{huang2026rzero} co-evolves a Challenger and Solver, rewarding the Challenger for proposing tasks near the edge of the Solver's capability. Question-begets-Question~\citep{bao2026question} instead repeatedly selects problems that the current model can mostly solve and uses a teacher to generate new variants for subsequent RL training. SPADE~\citep{liu2026spade} uses an LLM Environment Designer to generate executable environments and optimises generation using the Reasoning Agent's performance gap with and without privileged hints. These methods adapt task generation itself to the learner, typically through an LLM proposer, teacher, or environment designer.  Frontier learning instead searches an existing procedural generator's explicit configuration space, trading generation flexibility for inexpensive and interpretable exploration and mutation without an LLM-based proposal or rewriting stage.


\section{Procedural Generator}
\label{app:procedural_generator}

\begin{figure}[h]
  \centering
  \includegraphics[width=0.95\linewidth]{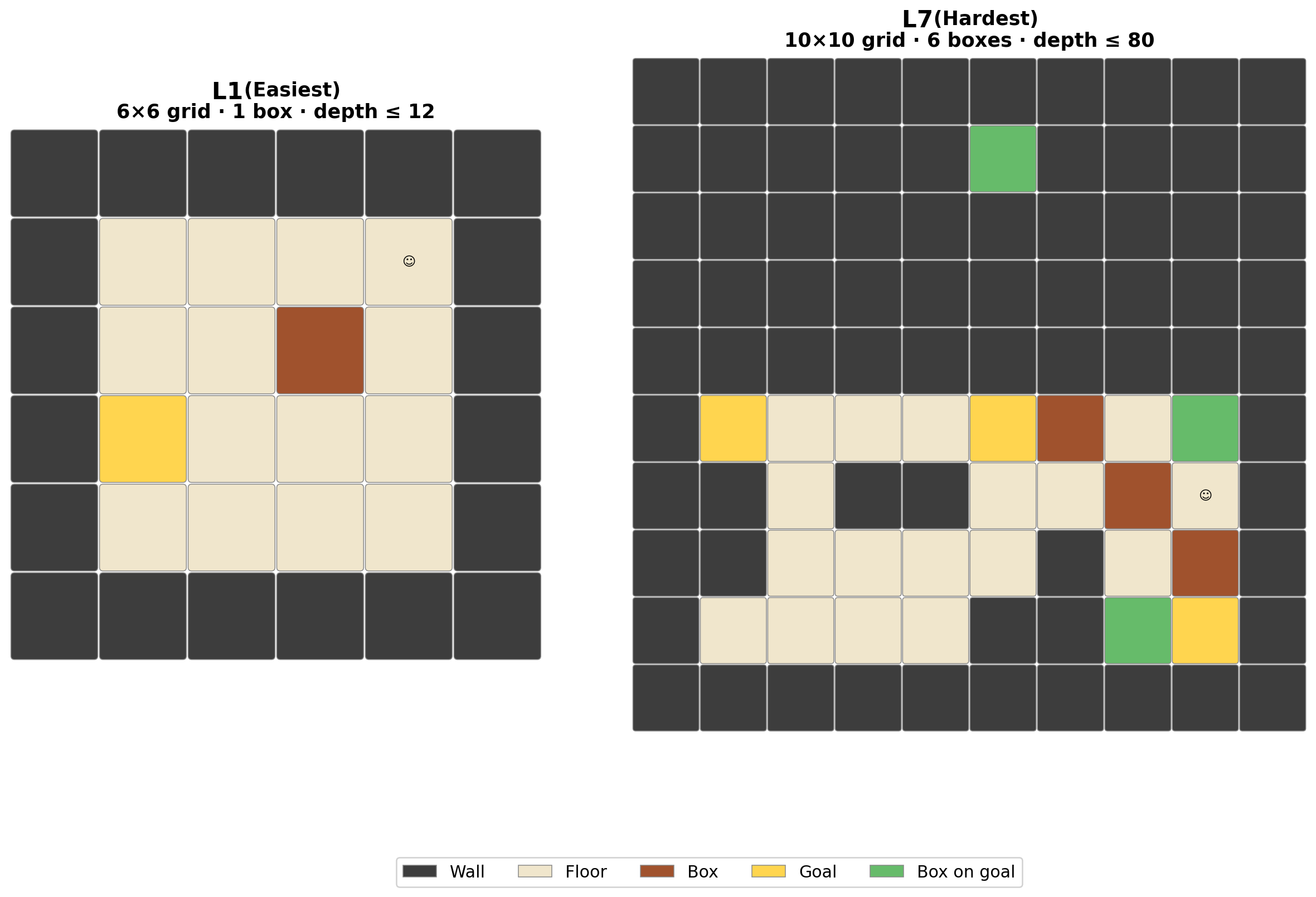}
  \caption{Example Sokoban levels at the two extremes of the anchor ladder. \textbf{Left:} L1 --- grid $6{\times}6$, 1 box, depth $\leq 12$. \textbf{Right:} L7 --- grid $10{\times}10$, 6 boxes, depth $\leq 80$. Colours: dark grey = wall, beige = floor, brown = box, yellow = goal, green = box on goal.}

  \label{fig:sokoban_levels}
\end{figure}

Figure~\ref{fig:sokoban_levels} illustrates how the Sokoban procedural generator produces levels of varying difficulty by scaling three configuration attributes: \textbf{grid size} (the side length of the square playable area), \textbf{number of boxes} to push onto goals, and \textbf{maximum solution depth} (the minimum number of moves required). 

\section{Anchor Levels Definition}
\label{app:anchor_levels}
Table~\ref{tab:anchor_levels} gives the full per-level specification for every task.
Countdown uses a monotone ladder co-scaling all attributes.
Sokoban levels cover distinct combinations of grid size, box count, and solution
depth without enforcing a strict monotone ordering on any single axis (difficulty
in Sokoban is non-monotone: a smaller grid with more boxes can be harder than a
larger one with fewer).
Dice is a Cartesian product over two independent axes.
Decimal Arithmetic jointly scales three axes.

\clearpage

\begin{table}[h!]
\centering
\caption{Full anchor-level specifications, grouped into Easy/Medium/Hard
(and Extra Hard for Dice) difficulty bins used throughout the paper.
Countdown: \texttt{num} = \texttt{num\_numbers},
\texttt{mv} = \texttt{max\_value}, \texttt{mt} = \texttt{max\_target}.
Sokoban: grid is square; \texttt{d} = \texttt{max\_depth}.
Decimal: \texttt{dp} = \texttt{decimal\_places}, \texttt{pr} = \texttt{precision}, \texttt{t} = \texttt{terms}. Largest Island: grid = \texttt{rows}$\times$\texttt{cols}; \texttt{isl} = \texttt{num\_islands}; \texttt{sz} = \texttt{island\_size}.}
\label{tab:anchor_levels}
\small
\setlength{\tabcolsep}{5pt}
\begin{tabular}{clcrrr}
\toprule
Task & Level & Bin & \multicolumn{3}{c}{Config} \\
\midrule
\multirow{10}{*}{Countdown}
 & 1  & Easy   & \texttt{num}=3   & \texttt{mv}=20  & \texttt{mt}=80   \\
 & 2  & Easy   & 3    & 40  & 150  \\
 & 3  & Easy   & 3--4 & 60  & 250  \\
 & 4  & Medium & 4    & 80  & 350  \\
 & 5  & Medium & 4--5 & 100 & 500  \\
 & 6  & Medium & 5    & 120 & 800  \\
 & 7  & Hard   & 5--6 & 150 & 1200 \\
 & 8  & Hard   & 6    & 180 & 2000 \\
 & 9  & Hard   & 6    & 220 & 3500 \\
 & 10 & Hard   & 6    & 300 & 5000 \\
\midrule
\multirow{7}{*}{Sokoban}
 & 1 & Easy   & grid=6  & boxes=1 & \texttt{d}12  \\
 & 2 & Easy   & 7  & 1 & 20  \\
 & 3 & Easy   & 8  & 2 & 30  \\
 & 4 & Medium & 9  & 2 & 45  \\
 & 5 & Medium & 8  & 3 & 45  \\
 & 6 & Hard   & 10 & 4 & 65  \\
 & 7 & Hard   & 10 & 6 & 80  \\
\midrule
\multirow{16}{*}{Dice}
 & 1  & Easy       & \texttt{num\_dice}=2 & \texttt{faces}=8  & \\
 & 2  & Easy       & 2 & 10 & \\
 & 3  & Easy       & 2 & 16 & \\
 & 4  & Easy       & 2 & 20 & \\
 & 5  & Medium     & 3 & 8  & \\
 & 6  & Medium     & 3 & 10 & \\
 & 7  & Medium     & 3 & 16 & \\
 & 8  & Medium     & 3 & 20 & \\
 & 9  & Hard       & 4 & 8  & \\
 & 10 & Hard       & 4 & 10 & \\
 & 11 & Hard       & 4 & 16 & \\
 & 12 & Hard       & 4 & 20 & \\
 & 13 & Extra Hard       & 5 & 8  & \\
 & 14 & Extra Hard & 5 & 10 & \\
 & 15 & Extra Hard & 5 & 16 & \\
 & 16 & Extra Hard & 5 & 20 & \\
\midrule
\multirow{9}{*}{Decimal Arith.}
 & 1 & Easy   & \texttt{dp}=4  & \texttt{pr}=8  & \texttt{t}=4  \\
 & 2 & Easy   & 4  & 12 & 6  \\
 & 3 & Easy   & 6  & 12 & 6  \\
 & 4 & Medium & 6  & 18 & 10 \\
 & 5 & Medium & 10 & 18 & 10 \\
 & 6 & Medium & 10 & 24 & 14 \\
 & 7 & Hard   & 14 & 18 & 10 \\
 & 8 & Hard   & 14 & 24 & 18 \\
 & 9 & Hard   & 18 & 24 & 14 \\
\midrule
\multirow{10}{*}{Largest Island}
 & 1  & Easy   & grid=8$\times$10  & \texttt{isl}=2 & \texttt{sz}=3 \\
 & 2  & Easy   & 9$\times$12       & 2 & 4 \\
 & 3  & Easy   & 10$\times$14      & 3 & 4 \\
 & 4  & Medium & 11$\times$15      & 3 & 5 \\
 & 5  & Medium & 12$\times$16      & 4 & 5 \\
 & 6  & Medium & 13$\times$17      & 4 & 6 \\
 & 7  & Hard   & 14$\times$18      & 5 & 6 \\
 & 8  & Hard   & 15$\times$18      & 5 & 7 \\
 & 9  & Hard   & 15$\times$18      & 6 & 6 \\
 & 10 & Hard   & 15$\times$18      & 6 & 7 \\
\bottomrule
\end{tabular}
\end{table}

\section{Implementation Details and Hyperparameters}
\label{app:hyperparameters}

Tables~\ref{tab:hp_grpo} and~\ref{tab:hp_accel} report the full set of
hyperparameters used in all experiments.
All runs using Qwen3-4B-Base or LLaMA3.2 set a maximum response length of 2048 tokens, while for \textsc{Sokoban} we use Qwen3-4B (thinking mode) with a response length of 4096 tokens to accommodate its extended chain-of-thought reasoning budget.
All methods are run with 4 random seeds (42, 123, 314, 999). All the runs are performed on H200 or GH200 GPUs, ensuring that experiments belonging to the same task family are run on the same GPU type.

\begin{table}[h]
\centering
\caption{GRPO training hyperparameters (shared across all methods and tasks).}
\label{tab:hp_grpo}
\small
\begin{tabular}{lcc}
\toprule
Hyperparameter & Symbol & Value \\
\midrule
Learning rate & & $10^{-6}$ \\
Train batch size & & 64 \\
Rollouts per problem & $n_r$ & 8 \\
PPO mini-batch size & & 16 \\
PPO micro-batch size per GPU & & 8 \\
Max problem length (tokens) & & 1024 \\
Max response length --- base model (tokens) & & 2048 \\
Max response length --- thinking model (tokens) & & 4096 \\
KL coefficient & $\beta_{\mathrm{KL}}$  & 0 | $10^{-4}$$^\dagger$ \\
Evaluation batch size & & 128 \\
Random seeds & & 4 \\
Clip low & $\epsilon_{low}$ & 0.2 \\
Clip high & $\epsilon_{high}$ & 0.2 | 0.28$^\dagger$ \\
\bottomrule
\end{tabular}
\\ {\footnotesize $^\dagger$A small KL loss and an asymmetric ("clip-higher") PPO
clipping range are applied for long-training stabilization only for the 500 steps runs (across all methods), preventing
entropy collapse over extended training horizons~\cite{liu2025prorlprolongedreinforcementlearning}.}
\end{table}

\begin{table}[h]
\centering
\caption{Frontier Learning curriculum hyperparameters.}
\label{tab:hp_accel}
\small
\begin{tabular}{lcc}
\toprule
Hyperparameter & Symbol & Value \\
\midrule
\multicolumn{3}{l}{\textit{Buffer and sampling}} \\
Levels per step & $n_\ell$ & 4 \\
Problems per level & $n_p$ & 16 \\
Buffer capacity & $B_{\max}$ & 100 \\
Seed levels & & 8 \\
Seed sampling mode & & grid \\
Sampling strategy & & Zipf \\
Zipf temperature & $T_z$ & 1.0 \\
Staleness coefficient & $\lambda_s$ & 0.05 \\
Informative success-rate interval & & [0.05-0.95] \\
\midrule
\multicolumn{3}{l}{\textit{Regret signal}} \\
Rolling window size & $n_w$ & 16 \\
Initial Regret & $R_0$ & 0.5 \\
\midrule
\multicolumn{3}{l}{\textit{Exploration}} \\
Explore fraction & $\phi$ & 0.3 \\
\midrule
\multicolumn{3}{l}{\textit{State-dependent mutation probabilities}} \\
Informative & $p_{\mathrm{inf}}$ & 0.4 \\
Too easy (mastered) & $p_{\mathrm{easy}}$ & 0.25 \\
Too difficult (impossible) & $p_{\mathrm{hard}}$ & 0.02 \\
Unseen (never sampled) & $p_{\mathrm{uns}}$ & 0.0 \\
\midrule
SEC (baseline) \\
EMA coefficient & $\alpha$ & 0.1 \\
\bottomrule
\end{tabular}
\end{table}
\clearpage

\section{Learning Curves }
\label{app:learning_curves}

\begin{figure}[h!]
\centering
\includegraphics[width=\linewidth]{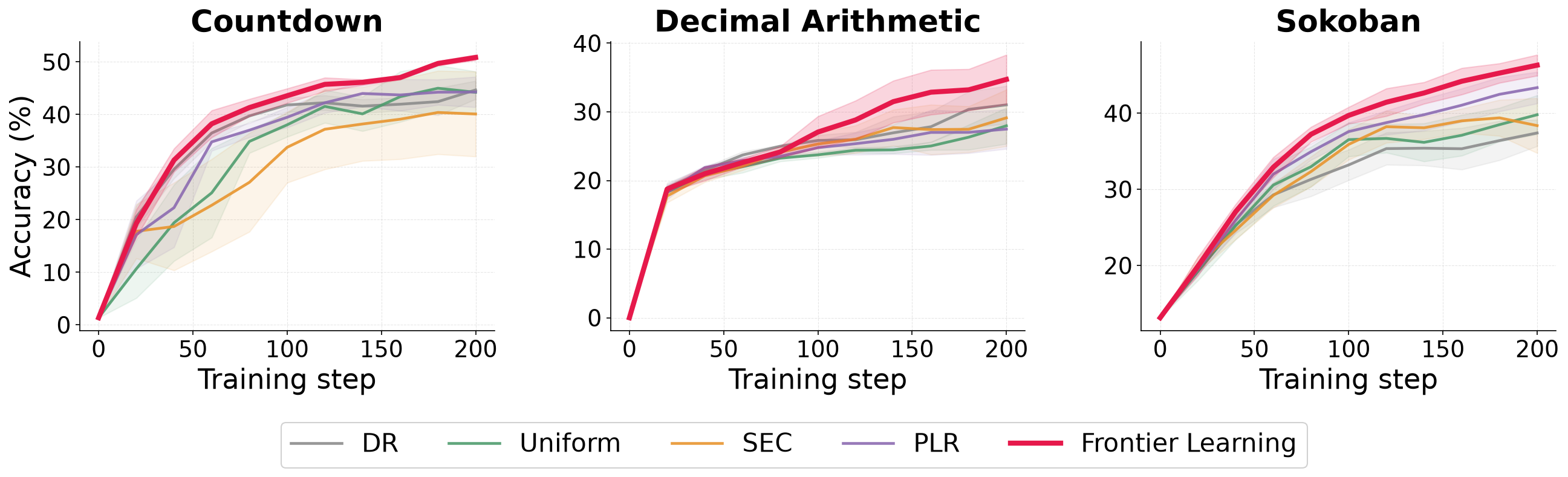}
\caption{
Accuracy (\%) on the fixed anchor evaluation sets over 200 GRPO steps for
\textsc{Countdown}, \textsc{Sokoban}, and \textsc{Decimal Arithmetic}.
}
\label{fig:learning_curves_standard}
\vspace{-1em}
\end{figure}

Figure~\ref{fig:learning_curves_standard} shows accuracy learning curves for Countdown, Decimal Arithmetic and Sokoban over the standard training budget of 200 GRPO steps.

\section{Per-Difficulty Accuracy Breakdown}

\label{app:difficulty_breakdown}

\begin{figure}[H]
    \centering
    \includegraphics[width=\linewidth]{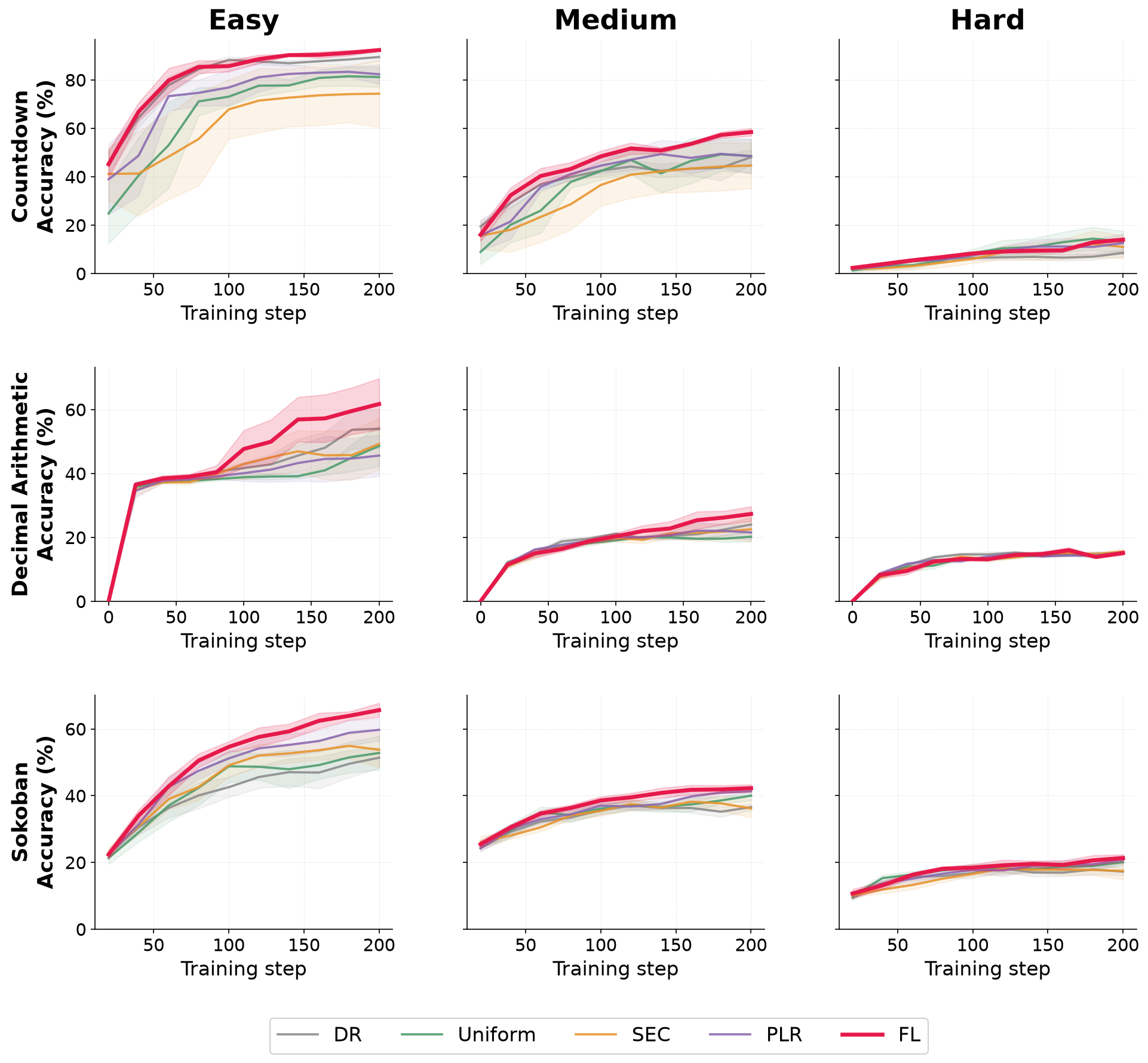}
    \caption{Validation accuracy across the three difficulty bins throughout the full training horizon for \textsc{Countdown}, \textsc{Sokoban}, and \textsc{Decimal Arithmetic}.}
    \label{fig:level_bin_curve_200_steps}
\end{figure}

\begin{figure}[H]
  \centering
  \includegraphics[width=\linewidth]{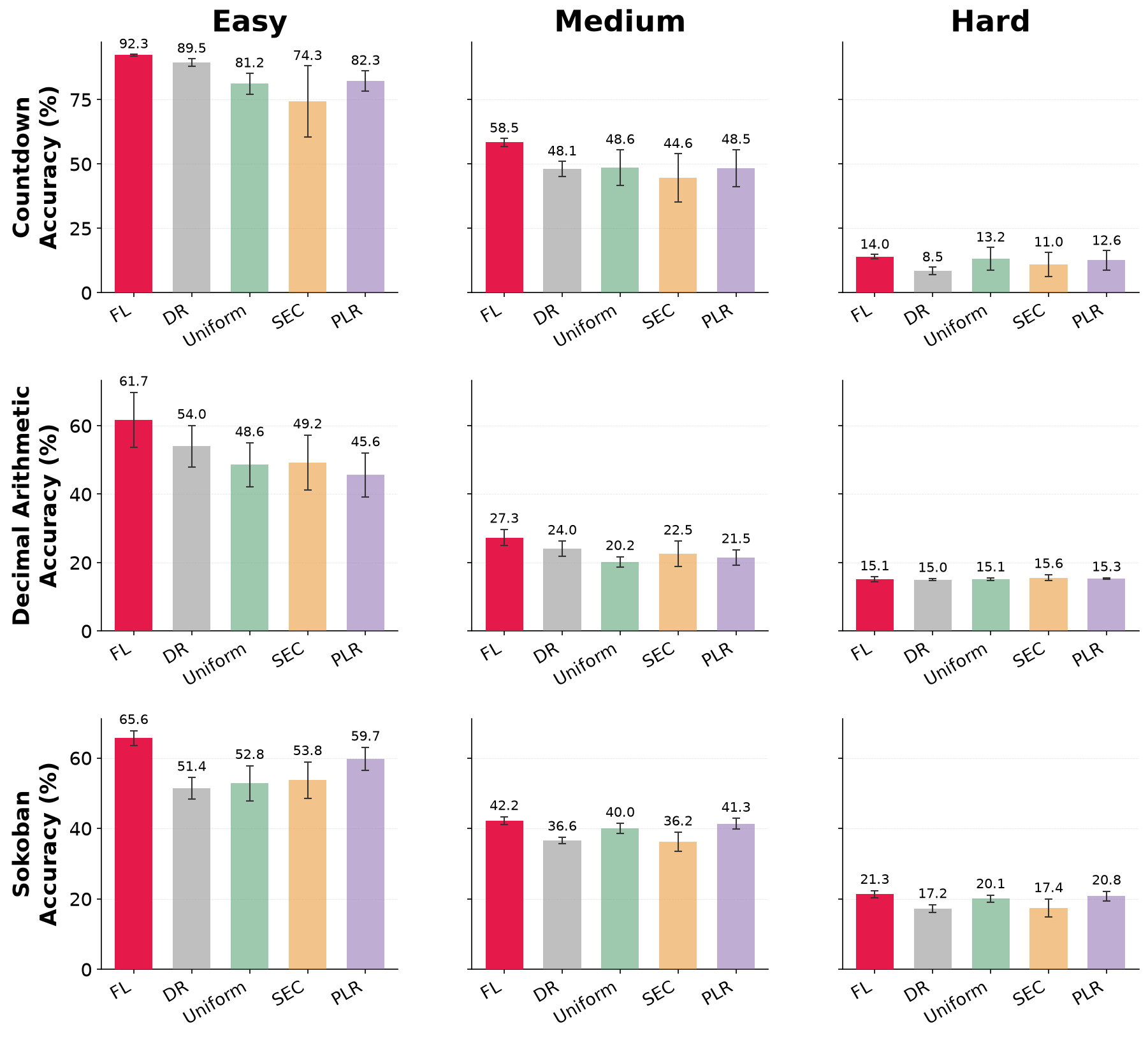}
  \caption{%
    Final validation accuracy per difficulty bin
    for \textsc{Countdown}, \textsc{Sokoban}, and \textsc{Decimal Arithmetic}.
  }
  \label{fig:level_bins_bars}
\end{figure}

Figure~\ref{fig:level_bin_curve_200_steps} reports the learning curves across the three tasks over 200 GRPO steps across the difficulties bins, while Figure~\ref{fig:level_bins_bars} breaks down the final validation performance by difficulty. On \textsc{Countdown} and \textsc{Decimal Arithmetic}, the gains from Frontier Learning are concentrated primarily on the Easy and Medium bins, whereas performance on the Hard bin remains largely comparable across methods. A similar pattern emerges on \textsc{Sokoban}, where Frontier Learning achieves its largest improvements on the easier configurations, with smaller gains as difficulty increases.

\begin{figure}[H]
    \centering
    \includegraphics[width=\linewidth]{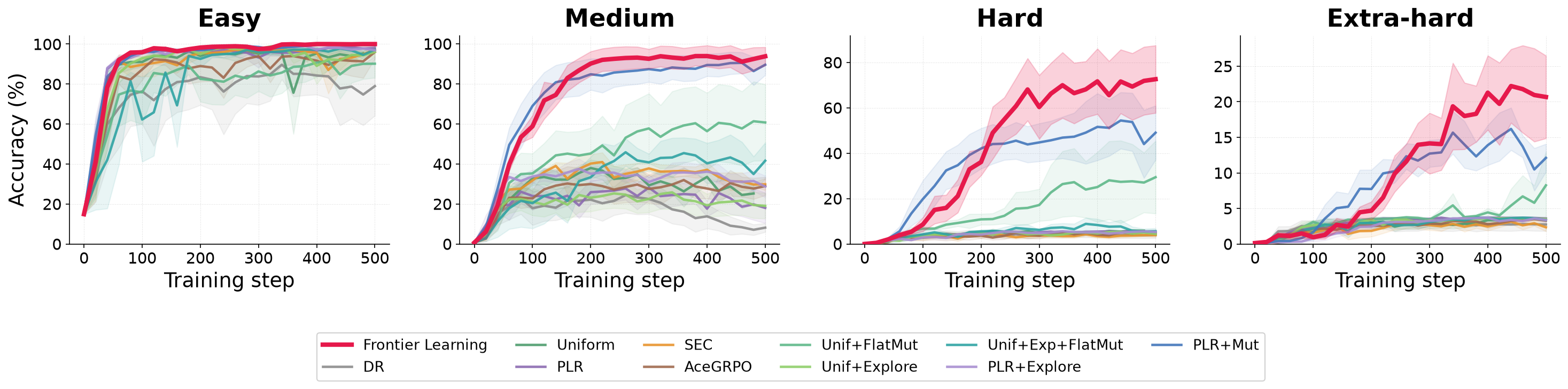}
    \caption{Validation accuracy across the four difficulty bins throughout the full training horizon for \textsc{Dice}.}
    \label{fig:level_bin_curve_dice}
\end{figure}

This behavior is consistent with the objective of Frontier Learning: prioritizing levels that lie near the model's current learning frontier, where additional training is most likely to yield improvement, rather than allocating substantial compute to levels that are either already mastered or currently too difficult to solve reliably. Importantly, this frontier is not fixed, but shifts as the model's capabilities improve. This is illustrated by the extended training results in Figure~\ref{fig:level_bin_curve_dice}, where Frontier Learning continues to improve throughout the 500-step training horizon, progressively reaching substantially higher validation accuracy on harder tasks. These results suggest that, as training progresses, the set of learnable configurations expands toward harder regions of the level space.


\begin{figure}[H]
\centering
\includegraphics[width=0.8\textwidth]{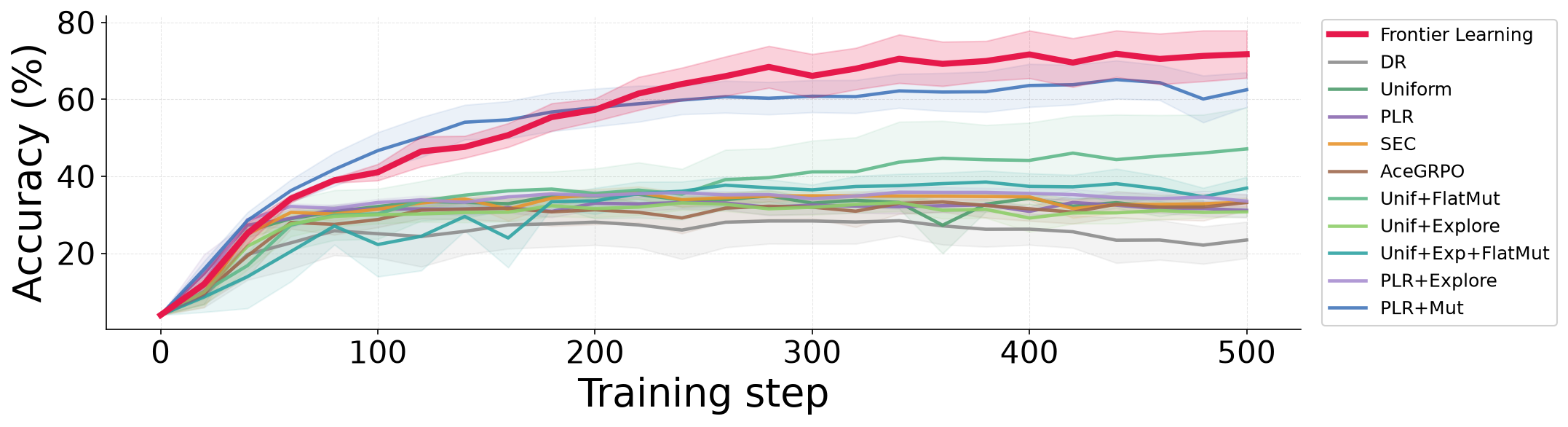}
\vspace{-1em}
\caption{
Long-horizon \textsc{Dice} learning curves over 500 GRPO steps. DAPO is omitted from the
step-based curves because it is matched by total rollout budget rather than
optimizer updates.
}
\label{fig:dice_learning_curves}
\end{figure}

\begin{figure}[H]
    \centering
    \includegraphics[width=0.9\linewidth]{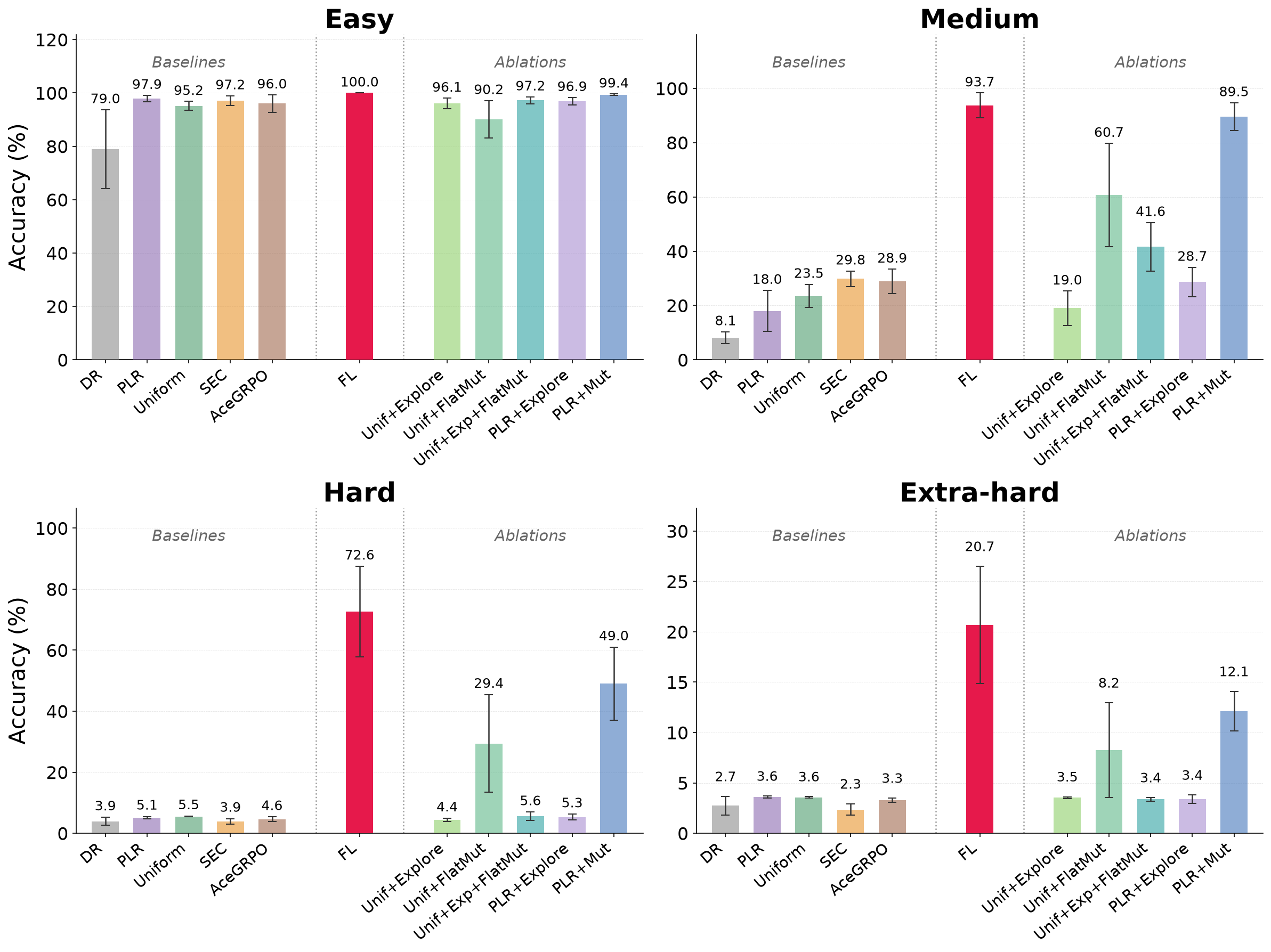}
    \caption{ Final validation accuracy per difficulty bin for \textsc{Dice}}
    \label{fig:dice-final-bin-difficulty}
\end{figure}

\section{Hyperparameter sensitivity}
\subsection{Frontier Learning}
\label{app:hyperparameter_sensitivity}

We additionally study the sensitivity of frontier learning to the staleness coefficient $\lambda_s$ and the informative success-rate interval defined in Equation \ref{eq:mutprob}. We conduct this ablation on the same Llama-3.2-3B-Instruct \textsc{Countdown} setting used for the cross-model-family evaluation in the main paper. Table~\ref{tab:llama_hparam_sensitivity} reports the results.

\begin{table}[h]
\centering
\caption{
Hyperparameter sensitivity on \textsc{Countdown} using Llama-3.2-3B-Instruct.
}
\label{tab:llama_hparam_sensitivity}
\setlength{\tabcolsep}{6pt}
\begin{tabular}{lc}
\toprule
Variant & Accuracy (\%) \\
\midrule
Default frontier learning                             & $39.2 \pm 1.3$ \\
$\lambda_s = 0.02$                     & $40.7 \pm 1.1$ \\
Informative interval $= [0.10, 0.90]$ & $40.5 \pm 0.8$ \\
Informative interval $= [0.02, 0.98]$ & $39.8 \pm 1.0$ \\
$\lambda_s = 0$                        & $38.8 \pm 1.4$ \\
\bottomrule
\end{tabular}
\end{table}

Performance remains within a narrow range across the tested configurations. In particular, varying either the staleness coefficient or the informative success-rate interval leads to only modest changes relative to the default configuration, suggesting that frontier learning is not highly sensitive to these hyperparameters around the chosen defaults.

\subsection{Step size sensitivity of the baselines}
\label{app:lr_sweep}
\begin{figure}[H]
    \centering
    \includegraphics[width=\linewidth]{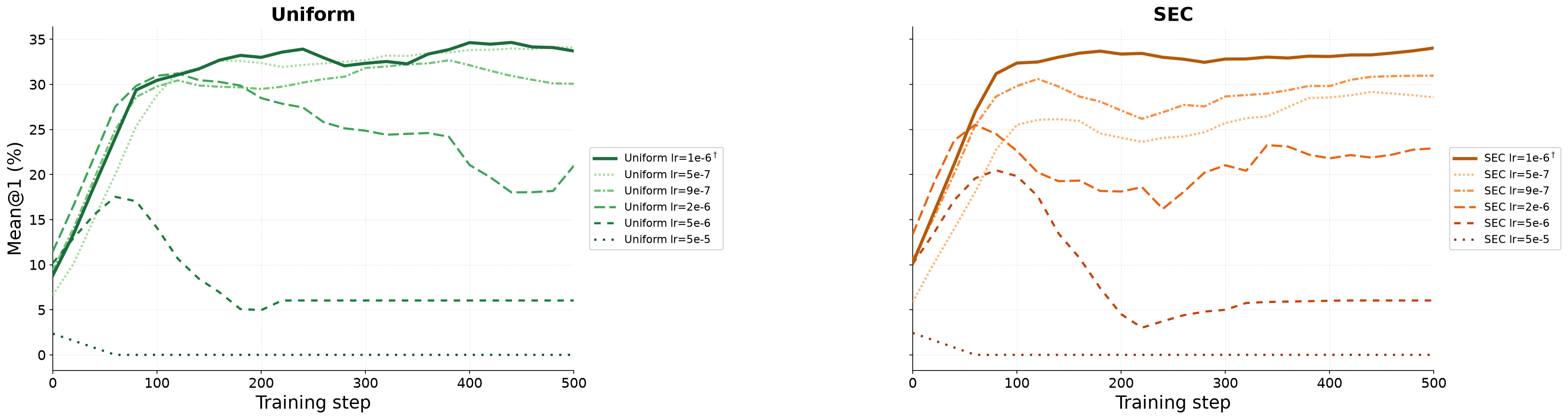}
    \caption{
    Learning-rate sweep on \textsc{Dice} (seed 42) for Uniform and
    SEC, the non-regret-based baseline and the best-performing baseline in
    Table~\ref{tab:dice_500}, respectively. 
    $^\dagger$Denotes the learning rate
    used for the main paper's results ($1\times10^{-6}$). No swept value raises
    either baseline's final accuracy above $\approx$35\%, indicating that an
    under-tuned learning rate does not explain the gap to frontier learning.
    }
    \label{fig:dice_lr_sweep}
\end{figure}

To rule out an under-tuned learning rate as an explanation for frontier
learning's gains, we swept the learning rate on \textsc{Dice} (500 steps,
seed 42) for two representative baselines: SEC, the best-performing baseline
in Table~\ref{tab:dice_500}, and Uniform, a non-regret-based method (fixed
pool, uniform sampling, no prioritization signal). We swept
$\{5\times10^{-7}, 9\times10^{-7}, 1\times10^{-6}, 2\times10^{-6},
5\times10^{-6}, 5\times10^{-5}\}$, holding every other hyperparameter fixed.
As it can be observed in Fig.~\ref{fig:dice_lr_sweep} our paper's learning rate, $1\times10^{-6}$, is the best- or
joint-best-performing value for both methods. Larger learning rates ($\geq 2\times10^{-6}$) are
uniformly worse and increasingly unstable, collapsing to near-zero accuracy
by $2\times10^{-6}$--$5\times10^{-5}$; no value we tested lifts either
baseline's final accuracy above $\approx$35\%, well below frontier
learning's performance under the same protocol.

\section{Compute accounting.}
\label{app:compute}

\paragraph{200-step experiments:}
We first report end-to-end compute for the 200-step experiments on
\textsc{Sokoban}, \textsc{Countdown}, and \textsc{Decimal Arithmetic}.
All methods use the same number of optimizer updates, prompts, rollouts per
prompt, and maximum response length. Across these tasks, frontier learning has
comparable wall-clock cost to the baselines and does not systematically generate
longer responses.

\begin{table}[h]
\centering
\caption{End-to-end compute accounting on Sokoban (200 steps).}
\label{tab:compute_sokoban}
\begin{tabular}{lccc}
\toprule
Method & Runtime (h) & Response length (tokens) & Accuracy (\%) \\
\midrule
Domain Rand. & 15.4$\pm$2.0 & 3838$\pm$33 & 37.6$\pm$2.6 \\
PLR          & 14.8$\pm$1.8 & 3698$\pm$103 & 41.1$\pm$3.9 \\
Uniform      & 15.1$\pm$1.8 & 3761$\pm$51 & 39.5$\pm$1.3 \\
SEC          & 14.8$\pm$1.8 & 3709$\pm$81 & 34.3$\pm$6.6 \\
Frontier Learning (ours)    & 14.2$\pm$1.7 & 3529$\pm$49 & 45.2$\pm$1.9 \\
\bottomrule
\end{tabular}
\end{table}

\begin{table}[h]
\centering
\caption{End-to-end compute accounting on Countdown (200 steps).}
\label{tab:compute_countdown}
\begin{tabular}{lccc}
\toprule
Method & Runtime (h) & Response length (tokens) & Accuracy (\%) \\
\midrule
Domain Rand. & 17.4$\pm$0.1 & 1538$\pm$14 & 44.7$\pm$1.7 \\
PLR          & 11.3$\pm$0.3 & 1521$\pm$44 & 44.3$\pm$2.9 \\
Uniform      & 11.8$\pm$0.3 & 1546$\pm$50 & 44.2$\pm$4.0 \\
SEC          & 11.2$\pm$0.4 & 1446$\pm$65 & 40.1$\pm$8.1 \\
Frontier Learning (ours)    & 11.7$\pm$0.5 & 1269$\pm$28 & 50.9$\pm$0.5 \\
\bottomrule
\end{tabular}
\end{table}

\begin{table}[H]
\centering
\caption{End-to-end compute accounting on Decimal Arithmetic (200 steps).}
\label{tab:compute_decimal}
\begin{tabular}{lccc}
\toprule
Method & Runtime (h) & Response length (tokens) & Accuracy (\%) \\
\midrule
Domain Rand. & 4.3$\pm$1.1 & 837$\pm$31 & 31.0$\pm$2.8 \\
PLR          & 6.0$\pm$1.1 & 677$\pm$30 & 27.5$\pm$2.9 \\
Uniform      & 5.7$\pm$1.0 & 642$\pm$33 & 28.0$\pm$2.6 \\
SEC          & 6.1$\pm$1.2 & 665$\pm$54 & 29.1$\pm$4.1 \\
Frontier Learning (ours)    & 4.2$\pm$1.1 & 758$\pm$50 & 34.7$\pm$3.6 \\
\bottomrule
\end{tabular}
\end{table}

\paragraph{Long-horizon compute on \textsc{Dice}:}
The 500-step \textsc{Dice} experiment reveals a different regime. Although all
methods except DAPO use the same number of optimizer updates, prompts, rollouts
per prompt, and maximum response length, realized wall-clock cost differs
substantially. Frontier learning requires $27.8\pm0.6$ hours, compared with
$18.7\pm1.1$ hours for PLR, and produces $1383\pm26$ tokens per response on
average, compared with $877\pm65$ for PLR.

This increase in response length is consistent with the outward movement of the
learning frontier observed in the per-difficulty analysis. After 500 steps, frontier learning
has nearly saturated Easy and Medium configurations and reaches 72.6\% on
Hard and 20.7\% on Extra-hard levels, whereas no standard baseline exceeds
5.5\% on these harder categories. As frontier learning acquires the ability to engage
with these more difficult configurations, its sampled problems induce longer
reasoning traces. The additional wall-clock cost is therefore partly a
consequence of the capability expansion produced during training, rather than
simply a larger nominal training budget.

However, this additional runtime does not explain frontier learning's performance
advantage. As shown in Table~\ref{tab:dice_matched_compute}, looking at the wall-clock time corresponding to PLR's complete 500-step run (i.e. the fastest baseline) frontier learning already reaches $68.0\pm5.5$\% accuracy.
Moreover, PLR+Mut produces even longer responses ($1426\pm14$ tokens) and
requires slightly more wall-clock time ($28.5\pm0.3$ hours) than frontier learning, while
attaining lower final performance. Thus, neither longer generations nor greater
realized wall-clock compute alone account for the gains of frontier learning.

\begin{table}[h]
\centering
\caption{End-to-end compute accounting on Dice (500 steps).}
\label{tab:compute_dice}
\begin{tabular}{lccc}
\toprule
Method & Runtime (h) & Response length (tokens) & Accuracy (\%) \\
\midrule
\multicolumn{4}{l}{\textit{Baselines}} \\
\addlinespace[2pt]
Domain Rand.      & 19.4$\pm$1.9 & 907$\pm$104  & 23.4$\pm$4.7 \\
PLR               & 18.7$\pm$1.1 & 877$\pm$65   & 31.2$\pm$1.9 \\
Uniform           & 20.8$\pm$1.9 & 994$\pm$108  & 31.9$\pm$1.5 \\
SEC               & 18.9$\pm$2.3 & 866$\pm$129  & 33.3$\pm$1.2 \\
ACE-GRPO          & 22.0$\pm$0.8 & 1063$\pm$45  & 33.2$\pm$2.1 \\
\midrule
\multicolumn{4}{l}{\textit{Component ablations}} \\
\addlinespace[2pt]
Unif+Explore      & 20.4$\pm$1.8 & 974$\pm$105  & 30.8$\pm$1.3 \\
Unif+FlatMut      & 22.6$\pm$1.3 & 1099$\pm$72  & 47.1$\pm$10.8 \\
Unif+Exp+FlatMut  & 20.1$\pm$1.8 & 941$\pm$110  & 36.9$\pm$2.9 \\
PLR+Explore       & 24.3$\pm$0.8 & 1190$\pm$45  & 33.5$\pm$1.8 \\
PLR+Mut           & 28.5$\pm$0.3 & 1426$\pm$14  & 62.5$\pm$4.5 \\
\midrule
frontier learning(ours)         & 27.8$\pm$0.6 & 1383$\pm$26  & 71.8$\pm$6.2 \\
\bottomrule
\end{tabular}
\end{table}

\begin{table}[h]
\centering
\caption{Dice performance at matched PLR compute time (18.7 hours).}
\label{tab:dice_matched_compute}
\begin{tabular}{lc}
\toprule
Method & Accuracy (\%) \\
\midrule
DR                & 25.4$\pm$3.7 \\
\textbf{PLR}      & \textbf{31.2$\pm$1.9} \\
Uniform           & 32.2$\pm$1.7 \\
SEC               & 33.5$\pm$1.6 \\
AceGRPO           & 31.8$\pm$3.3 \\
\textbf{Frontier Learning (ours)}       & \textbf{68.0$\pm$5.5} \\
Unif+FlatMut      & 43.7$\pm$9.8 \\
Unif+Explore      & 31.9$\pm$1.7 \\
Unif+Exp+FlatMut  & 37.1$\pm$3.4 \\
PLR+Explore       & 34.8$\pm$1.2 \\
PLR+Mut           & 60.7$\pm$4.3 \\
\bottomrule
\end{tabular}
\end{table}

\paragraph{Long horizon compute on \textsc{Largest Island}:}
Unlike on \textsc{Dice}, frontier learning's runtime and response length on \textsc{Largest Island} are not elevated relative to the baselines: both sit in the middle of the
range they span (Table~\ref{tab:compute_largest_island}), so, also in this case, frontier learning's accuracy advantage cannot be attributed to spending more wall-clock time or producing
longer responses.

\begin{table}[h]
\centering
\caption{End-to-end compute accounting on Largest Island (500 steps, Olmo-3-7B-Instruct).}
\label{tab:compute_largest_island}
\begin{tabular}{lccc}
\toprule
Method & Runtime (h) & Response length (tokens) & Accuracy (\%) \\
\midrule
Domain Rand. & 63.2$\pm$0.3 & 1578$\pm$29 & 45.2$\pm$1.8 \\
PLR          & 44.3$\pm$6.0 & 1276$\pm$96 & 52.0$\pm$5.3 \\
Uniform      & 53.1$\pm$2.5 & 1132$\pm$125 & 45.3$\pm$3.4 \\
SEC          & 44.4$\pm$7.1 & 1218$\pm$102 & 48.3$\pm$4.7 \\
ACE-GRPO     & 43.8$\pm$4.7 & 1243$\pm$106 & 45.3$\pm$5.5 \\
frontier learning (ours) & 52.4$\pm$0.7 & 1498$\pm$6 & 73.7$\pm$3.3 \\
\bottomrule
\end{tabular}
\end{table}

\end{document}